\documentclass[12pt]{article}
\usepackage{amsmath,amssymb,amsthm,amsfonts}
\usepackage{arxiv}
\usepackage{hyperref}
\usepackage{nicefrac}
\usepackage{bbm}
\usepackage{mathrsfs}
\usepackage{algorithm}
\usepackage{algorithmic}
\usepackage{times}
\usepackage{graphicx}
\usepackage{subcaption}
\usepackage{url}
\usepackage{booktabs}
\usepackage{secdot}

\newtheorem{theorem}{Theorem}
\newtheorem{assumption}[theorem]{Assumption}
\newtheorem{remark}[theorem]{Remark}
\newtheorem{definition}[theorem]{Definition}
\newtheorem{corollary}[theorem]{Corollary}

\newcommand{\algname}[1]{{\sf  #1}}

\renewcommand{\endproof}{\hfill$\square$}

\newcommand{\1}{\mathbbm 1}

\newcommand{\R}{\mathbb{R}}

\def\<#1,#2>{\left\langle #1,#2\right\rangle}
\renewcommand{\leq}{\leqslant}

\renewcommand{\geq}{\geqslant}
\renewcommand{\ge}{\geqslant}

\newcommand{\Med}{\mathtt{Med}}
\newcommand{\Mean}{\mathtt{Mean}}
\newcommand{\SMM}{\mathtt{SMoM}}

\newcommand{\cG}{{\cal G}}

\newcommand{\cN}{{\cal N}}
\newcommand{\cO}{{\cal O}}

\newcommand{\mI}{{\bf I}}

\newcommand{\clip}{\texttt{clip}}

\newcommand{\EE}{\mathbb{E}}

\title{Fast UCB-type algorithms for stochastic bandits with heavy and super heavy symmetric noise}
\usepackage{times}

\author{
Yuriy Dorn\thanks{Corresponding author} \\
Institute for Artificial Intelligence \\ Lomonosov Moscow State University, Moscow, Russia\\
Moscow Institute of Physics and Technology, Moscow, Russia\\
\url{dornyv@my.msu.ru}
\And
Aleksandr Katrutsa \\
Skoltech, Moscow, Russia\\
AIRI, Moscow, Russia\\
\url{amkatrutsa@gmail.com} 
\AND
Ilgam Latypov\\
Institute for Artificial Intelligence \\ Lomonosov Moscow State University, Moscow, Russia\\
Moscow Institute of Physics and Technology, Moscow, Russia \\
\url{latypov.im@phystech.edu}
\And
Andrey Pudovikov\\
Institute for Artificial Intelligence\\ Lomonosov Moscow State University, Moscow, Russia\\
\url{pudovikov123@mail.ru}
}

\begin{document}

\maketitle

\begin{abstract}%
  In this study, we propose a new method for constructing UCB-type algorithms for stochastic multi-armed bandits based on general convex optimization methods with an inexact oracle. 
  We derive the regret bounds corresponding to the convergence rates of the optimization methods. 
  We propose a new algorithm \algname{Clipped-SGD-UCB} and show, both theoretically and empirically, that in the case of symmetric noise in the reward, we can achieve an $O(\log T\sqrt{KT\log T})$ regret bound instead of $O\left (T^{\frac{1}{1+\alpha}} K^{\frac{\alpha}{1+\alpha}} \right)$ for the case when the reward distribution satisfies $\mathbb{E}_{X \in \mathcal{D}}[|X|^{1+\alpha}] \leq \sigma^{1+\alpha}$ ($\alpha \in (0, 1])$, i.e. perform better than it is assumed by the general lower bound for bandits with heavy-tails. 
  Moreover, the same bound holds even when the reward distribution does not have the expectation, that is, when $\alpha<0$. 
\end{abstract}


\section{Introduction}

In this work, we consider the stochastic multi-armed bandit problem (MAB) with a heavy-tailed reward distribution introduced by~\cite{bubeck2013bandits}.
This problem is a special case of the classical MAB problem introduced by~\cite{robbins1952some} with a lower bound  $\Omega\left ( T^{\frac{1}{1+\alpha}} K^{\frac{\alpha}{1+\alpha}} \right )$. 
The problem is formulated as follows: an agent sequentially chooses one of the $K$ actions (arms) in every round with a total number of rounds equal to $T$. 
For each arm $i=1, \dots, K$ there is a corresponding unknown probability distribution $\mathcal{D}_i$ with a finite mean $\mu_i$ and finite $(1+\alpha)$-moment with $\alpha \in (0, 1]$. 
That is, for each arm $i$, there exists fixed $\sigma_i>0$ such that $\EE_{X \in \mathcal{D}_i}[|X|^{1+\alpha}] \leq \sigma_i^{1+\alpha}$. 
In each round $t$ when the agent picks an arm $A_t$ the reward is drawn independently from $\mathcal{D}_{A_t}$.

An agent aims to minimize the regret, accumulated throughout $T$ rounds
\[
R_T = T \max_{1\leq i \leq K} \mu_i - \sum_{t=1}^T \EE [\mu_{A_t}].
\]
The Robust UCB proposed in~\cite{bubeck2013bandits} is a general template for constructing UCB-type algorithms if one has an accessible and robust mean estimation procedure. 
In short, it can be described as follows:
\begin{itemize}
    \item Find mean estimation procedure with high probability deviation bound, i.e. for each arm $i$ find mean estimator $\hat{\mu}_i^{k_i}$ based on $k_i$ samples of reward, such that $|\hat{\mu}_i^{k_i} - \mu_i| \leq r(k_i, \delta)$ holds with probability at least $1-\delta$ and the confidence radius $r(k_i, \delta)$ degenerates over $k_i$ as fast as possible,
    \item For each arm $i$ construct upper confidence bound (UCB)
    $$
    UCB_i(k, \delta) = \hat{\mu}_i^k + r(k, \delta),
    $$ 
    which is used as a high probability upper bound on mean reward.
    \item Play arm $A_t$ with the highest current UCB $A_t = \arg \max_{1\leq i \leq K} UCB_i(k_i, \delta)$, receive feedback and update UCB estimation for the played arm.
\end{itemize}

In the vanilla UCB introduced by~\cite{auer2002finite}, this procedure uses the empirical mean $\hat{\mu}^{k_i}_i = \frac{1}{k_i} \sum_{t=1}^T X_t \1_{\{A_t = i\}}$ and confidence interval $r(k_i, \delta) = \sqrt{\frac{2v \log 1/\delta}{k_i}}$. 
In Robust UCB, the empirical mean is replaced by a truncated empirical mean, median of means, or Catoni’s M-estimator to construct $\hat{\mu}_i^{k_i}$ and confidence radius $r(k_i, \delta) = v^{1/\alpha} \left ( \frac{c \log 1/\delta}{k_i} \right )^{\frac{\alpha}{1+\alpha}}$ with an appropriate choice of parameters $v$ and $c$.

Index-based algorithms, such as UCB-type algorithms, are computationally intensive procedures. 
This reduces the practical usefulness of Robust UCB based on the truncated mean as well as recently proposed algorithms in~\cite{agrawal2021regret,genalti2023towards}. 
From the practical point of view ``optimality'' is the trade-off between iteration complexity and oracle complexity with convergence measured in seconds (not number of iterations) as a final judge.


\subsection{Related works}


The Robust UCB algorithm from~\cite{bubeck2013bandits} is probably the first relevant UCB-type algorithm for bandits with heavy tails. 
Study~\cite{agrawal2021regret} proposed optimal in the first-factor algorithm with faster concentration compared to well-known truncated or trimmed empirical mean estimators for the mean of heavy-tailed distributions.
The downside is that the proposed index is computationally demanding. 
The authors of~\cite{lee2020optimal,lee2022minimax} propose an optimal algorithm for the case when it is known that reward distribution has $p$-th moment, but the bounding constant is unknown. 
Adaptive Robust UCB algorithm~\cite{genalti2023towards} matches the lower bound and does not require any additional knowledge of the reward's distribution parameters. 

Study~\cite{lu2019optimal} considers Lipschitz bandits with heavy-tailed rewards and establishes corresponding lower bounds. 
The authors of~\cite{medina2016no} consider linear bandits with heavy-tailed rewards. 
Optimal algorithms for this setting (up to a logarithmic factor) were introduced in~\cite{shao2018almost}.


In recent years, another idea, usually referred to as "best-of-two-worlds,” was proposed. 
The name implies that the proposed algorithms achieve lower bounds in both stochastic and adversarial MAB settings. 
This idea assumes the application of solution techniques for adversarial bandits from~\cite{auer2002nonstochastic} to stochastic MAB problems with a heavy-tailed distribution of rewards, see~\cite{zimmert2019optimal, lee2020optimal, huang2022adaptive, zhang2022parameter, dorn2024implicitly}. 
Powered by recent advances in online convex optimization (OCO) (see~\cite{orabona2019modern}), this approach leads to optimal algorithms in both adversarial and stochastic settings with instance-independent regret bound $O\left ( T^{\frac{1}{1+\alpha}} K^{\frac{\alpha}{1+\alpha}} \right )$. 

Despite optimality, there exist cases in which the "best-of-the-two-worlds" framework is not the best choice. 
Because of the nature of the adversarial bandit problem, any reasonable solution includes additional randomization, and at each step, the algorithm proposes a probability distribution over arms to sample and adjusts it based on feedback. 
In the best case, as the rounds continue, the probability of picking the best arm converges to 1. 
In risk-sensitive applications, additional randomization can be problematic. 
Another point is that the probability distribution adjustment is done by online mirror descent and can sometimes be computationally intensive, for example, when it is difficult to compute the projections involved. 
Thus, these features of the "best-of-the-two-worlds" framework  restrict their usefulness for real-time risk-sensitive and highly loaded systems. 

\subsection{Contributions}
In this work, we follow the same UCB template but make it even more general. 
Most estimators can be considered as a solution to the corresponding optimization problem, as MLE is a maximizer of likelihood. 
Therefore, an estimator of the mean can be a solution to some stochastic optimization problem (with the expectation being a solution in the absence of noise). 
Most optimization methods are iterative and produce a sequence of estimators $\{x^k\}_{k=1}^{\infty}$ that converge to the optimal solution $\lim_{k \to \infty}x^k = x^*$. 
Good methods also provide convergence guarantees, sometimes with deviation bounds. 
Therefore, the natural idea is to construct $K$ auxiliary optimization problems for reward estimation, one for each arm, and use the appropriate (w.r.t. to accessible inexact oracle) optimization method and the corresponding convergence rate to construct the UCB-type index.

To proceed with this idea, we consider a stochastic bandit problem with a reward distribution that satisfies the following assumption, which is more convenient than $\EE_{X \in \mathcal{D}}[|X|^{1+\alpha}] \leq \sigma^{1+\alpha}$ ($\alpha \in (0, 1])$, particularly when the expected reward is not properly defined.

\begin{assumption}
    \label{as:convolution}
    For any arm $i$ ($i=1,\dots,K$) and any round $t$ ($t=1, \dots, T$), random reward $X_i^t = \mu_i + \xi_i^t$ and the probability density function $\rho_i^t$ of the noise $\xi_i^t$ satisfy the following condition:
    \begin{itemize}
    \item $\rho_i^t(u) = \rho_i(u)$ for any $u \in \R$ 
 (i.i.d.),
    \item $\rho(u) = \rho(-u)$ for any $u \in \R$  (symmetric noise),
    \item there are $\sigma > 0$ and $\alpha>0$, such that $\EE[|\xi_i^t|^{\alpha}] \leq \sigma^{\alpha}$ (heavy tail).
    \end{itemize}
\end{assumption}

To make a smooth transition from general optimization methods to UCB-type algorithms, we introduce the concept of $g(k, \delta)$-bounded algorithms in Subsection~\ref{sec: bounded_alg}. 
In Subsection~\ref{sec:FO/ZO} we propose a few appropriate auxiliary optimization problems for arm expected reward estimation, followed by the introduction of \algname{FO-UCB} and \algname{ZO-UCB} algorithms based on general $g(k, \delta)$-bounded first-order and zero-order algorithms for convex optimization with inexact oracles.

In Section~\ref{sec: convergence1} we prove regret bound for \algname{FO-UCB}
\begin{align}
    &R_T \leq \sum_{i: \Delta_i>0} \Delta_i \left (g^{-1}\left (\frac{\Delta_i^2}{8}, \frac{1}{T^2} \right ) + 2 \right )
\end{align}
and in Section~\ref{sec: convergence2} we prove the similar bound for the \algname{ZO-UCB}
$$
R_T = \sum_{i=1}^K \Delta_i \left ( g^{-1}\left (\frac{\Delta_i}{2}, \frac{1}{T^2} \right ) + 2\right ).
$$
In Section~\ref{sec:SGD}, based on convergence results for \algname{Clipped-SGD} from~\cite{puchkin2023breaking}, we prove that \algname{Clipped-SGD} is $g(k, \delta)$-bounded algorithm. 
Based on this result, we propose the \algname{FO-UCB}-type algorithm \algname{Clipped-SGD-UCB}  with regret $O(\log T\sqrt{KT\log T})$ for noise that satisfies Assumption~\ref{as:convolution}.
In Section~\ref{sec:NumExp} we show experimentally that our algorithm outperforms other UCB-type algorithms with heavy-tail and super-heavy-tail (no expectation) settings and performs nearly optimally in the case of sub-Gaussian noise.

In this work, we consider the case of symmetric noise, for which we obtain the regret bound $O(\log T\sqrt{TK\log T})$ even in a setting where the noise has no expectation. 
Because the Robust UCB with a median of means estimator is very close to that of our algorithm, it shows results close to ours. 
The only drawback is that Robust UCB is much more computationally demanding than our algorithm.

\section{UCB via stochastic optimization algorithms}

First, we introduce the concept of $g(k, \delta)$-bounded algorithms to clarify our analysis.

\subsection{Optimization  methods with inexact oracle}
\label{sec: bounded_alg}

For our work, we consider an unconstrained smooth convex optimization problem
\begin{equation*}
    \min_{x \in \R} f(x),
\end{equation*}
where the function $f: \R \to \R$ is accessible through the stochastic first-order oracle $\mathcal{G}: \R \to \R$ or via a stochastic zero-order oracle $\mathcal{H}: \R \to \R$. Denote  $x^* = \arg \min_{x \in \R} f(x)$

An algorithm
\[
x_{k+1} = \mathcal{A}\left (x_0, \mathcal{G}(x_0), \dots, x_k, \mathcal{G}(x_k) \right )
\]
is called $g(k, \delta)$-bounding first-order algorithm if for any $k \in \mathbb{N}$ and $\delta>0$ inequality
\[
f(x_k) - f(x^*) \leq g(k, \delta)
\]
holds with a probability of at least $1-\delta$.

An algorithm
$$
x_{k+1} = \mathcal{A}\left (x_0, \mathcal{H}(x_0), \dots, x_k, \mathcal{H}(x_k) \right )
$$
is called $g(k, \delta)$-bounding zero-order algorithm if for any $k \in \mathbb{N}$ and $\delta>0$ inequality
$$
f(x_k) - f(x^*) \leq g(k, \delta)
$$
holds with probability at least $1-\delta$.

\subsection{\algname{FO-UCB} and \algname{ZO-UCB} algorithms}
\label{sec:FO/ZO}

We now demonstrate how to construct UCB-type algorithms by incorporating $g(k, \delta)$-bounding optimization algorithms in the UCB framework.
\begin{enumerate}
    \item For each arm $i$ ($i=1, \dots, K$) construct supplementary convex optimization problem
    $$
    \min_{x \in \R}f_i(x),  
    $$
    such that $\mu_i = \arg \min_{x \in \R} f_i(x)$, with an accessible stochastic zero/first-order oracle and the corresponding $g(k, \delta)$-bounding algorithm $\mathcal{A}$. 
    \item Use $\{x^k_i\}_{k=1}^T$ generated by the algorithm $\mathcal{A}$ and corresponding bound $g(k, \delta)$ to construct UCB-type estimation on the mean $\mu_i$.
    \item At each round play arm with the biggest UCB.
\end{enumerate}

Setup examples:
\begin{itemize}
    \item [\textbf{\algname{FO-UCB}}] Let $f_i(x) = \frac{1}{2}(x-\mu_i)^2$ with admissible stochastic first order oracle $\mathcal{G}(x) = \nabla f_i (x, \xi) = x - \xi$. Suppose that we have $g(k, \delta)$-bounded first-order algorithm $\mathcal{A}$ for this problem, and the sequence $\{x_k\}_{k=1}^{\infty}$ is generated by $\mathcal{A}$. Then we propose the following index
    $$
    UCB(i, n_i(t), \delta) = x_i^{n_i(t)} + \sqrt{2g(n_i(t), \delta)},
    $$
    where $n_i(t) = \sum_{s=1}^t \1_{\{A_s = i\}}$ is the number of times the $i$th arm is chosen in the first $t$ rounds.
    \item [\textbf{\algname{ZO-UCB}}] Let $f_i(x) = |x-\mu_i|$ with admissible zero-order oracle $\mathcal{H}(x) = f_i(x, \xi) = |x - \xi|$. Suppose that we have $g(k, \delta)$-bounded zero-order algorithm $\mathcal{A}$ for this problem, and the sequence $\{x_k\}_{k=1}^{\infty}$ is generated by $\mathcal{A}$. Then we propose the following index
    $$
    UCB(i, n_i(t), \delta) = x_i^{n_i(t)} + g(n_i(t), \delta).
    $$
\end{itemize}

\subsection{Convergence of \algname{FO-UCB}}
\label{sec: convergence1}

\begin{theorem}[Convergence of \algname{FO-UCB}]\label{thm:FO_UCB} 
 The regret of the \algname{FO-UCB} with $g(k, \delta)$-bounded first-order algorithm for the MAB problem with $K$ arms, auxiliary functions $f_i(x) = \frac{1}{2}(x-\mu_i)^2$, period $T$, $\delta=\frac{1}{T^2}$ satisfies
\begin{align}
    &R_T \leq \sum_{i: \Delta_i>0} \Delta_i \left (g^{-1}\left (\frac{\Delta_i^2}{8}, \frac{1}{T^2} \right ) + 2 \right )
\end{align}
\end{theorem}
The proof of this theorem is given in Appendix~\ref{sec::proof-FO-UCB}.

\subsection{Convergence of \algname{ZO-UCB}}
\label{sec: convergence2}

\begin{theorem}[Convergence of \algname{ZO-UCB}]\label{thm:ZO_UCB} 
 The regret of the \algname{ZO-UCB} with $g(k, \delta)$-bounded first-order algorithm for the MAB problem with $K$ arms, auxiliary functions $f_i(x) = |x-\mu_i|$, period $T$, $\delta=\frac{1}{T^2}$ satisfies
$$
R_T = \sum_{i=1}^K \Delta_i \left ( g^{-1}\left (\frac{\Delta_i}{2}, \frac{1}{T^2} \right ) + 2\right )
$$
\end{theorem}
The proof of this theorem is given in Appendix~\ref{sec::proof-ZO-UCB}.

{\remark Note that these results do not mean that \algname{ZO-UCB} achieves better regret compared to \algname{FO-UCB} because bounding functions $g(k, \delta)$ for first-order and zero-order algorithms are different.}

Thus, we have obtained the general results for arbitrary $g(k, \delta)$-bounded algorithms. 
The next step is to proceed with a particular choice of the \algname{ZO-UCB} or \algname{FO-UCB} algorithm and show that these frameworks allow us to obtain very good variants of the UCB algorithms.

\subsection{\algname{Clipped-SGD-UCB}}
\label{sec:SGD}

As we can see from Theorem~\ref{thm:FO_UCB}, to make a good UCB-type algorithm we need a good (in the sense of bounding function $g(k,\delta)$, i.e. convergence) first-order algorithm. 
We proceed with \algname{clipped-SGD} algorithm.

\subsubsection{\algname{Clipped-SGD}}

Next, we use results obtained in~\cite{puchkin2023breaking} for \algname{clipped-SGD} with a smooth median of means as a gradient estimator.
In particular, we show that \algname{clipped-SGD} algorithm is $g(k, \delta)$-bounded first order algorithm and presents a particular form of the function $g$.
First, we proceed with the definition of smooth median of means.

\begin{definition}
    \label{def:smom}
    Let $\zeta$ be a random element in $\R^d$ and let $\theta > 0$ be an arbitrary number. For any positive integers $m$ and $n$, the smoothed median of means $\SMM_{m, n}(\zeta, \theta)$ is defined as follows:
    \begin{equation}
        \SMM_{m, n}(\zeta, \theta)
        = \Med\left(\upsilon_1, \dots, \upsilon_{2m +1}\right),
        \label{eq::smom_def}
    \end{equation}
    where, for each $j \in \{0, \dots, 2m\}$,
    \[
        \upsilon_j = \Mean(\zeta_{j n + 1}, \dots, \zeta_{(j + 1) n}) + \theta \, \eta_{j + 1},
    \]
    $\zeta_1, \dots, \zeta_{(2m + 1) n}$ are i.i.d. copies of $\zeta$, and
    $\eta_1, \dots, \eta_{2m + 1} \sim \cN(0, \mI_d)$ are independent standard Gaussian random vectors.
\end{definition}

\begin{assumption}
\label{as:bounded_bias_and_variance}
    There exists $N \in \mathbb{N}$, aggregation rule $\mathcal{R}$ and (possibly dependent on $T$) constants $b\geq 0$, $\sigma \geq 0$ such that for an $x \in \R$ i.i.d.\ samples $\nabla f_{\xi_1}(x),\ldots, \nabla f_{\xi_N}(x)$ from the oracle $\cG(x)$ satisfy the following relations:
    \[
\left|\EE[\nabla f_{\Xi}(x)] - \nabla f(x)\right| \leq b \quad \EE\left[\left\|\nabla f_{\Xi}(x) - \EE[\nabla f_{\Xi}(x)]\right\|^2\right] \leq \sigma^2, 
    \]
    where $\nabla f_{\Xi}(x) = \mathcal{R}(\nabla f_{\xi_1}(x),\ldots, \nabla f_{\xi_N}(x))$ and expectations are taken w.r.t.\ $\nabla f_{\xi_1}(x),\ldots, \nabla f_{\xi_N}(x)$.
\end{assumption}

Then \algname{clipped-SGD} algorithm can be defined as
\begin{equation}
    x^{k+1} = x^k - \gamma_k \clip(\nabla f_{\Xi^k}(x^k), \lambda_k), \label{eq:clipped_SGD}
\end{equation}
where $\nabla f_{\Xi^k}(x^k)$ is an estimator satisfying Assumption~\ref{as:bounded_bias_and_variance} sampled independently from previous iterations.
We also need the following assumptions for technical reasons.

\begin{assumption}
\label{as:L_smoothness}
    There exists a set $Q\subseteq \R^d$ and constant $L > 0$ such that for all $x, y \in Q$
    \[
    \|\nabla f(x) - \nabla f(y)\| \leq L\|x - y\|, \quad \|\nabla f(x)\|^2 \leq 2L\left(f(x) - f_*\right),
    \]
    where $f_* = \inf_{x \in Q}f(x) > -\infty$.
\end{assumption}


Now we are ready to present the particular case of a theorem from~\cite{puchkin2023breaking} to show that \algname{clipped-SGD} can be considered as an example of $g(k, \delta)$-bounding algorithm.

\begin{theorem}
\label{th::clip-sgd-th}
    Consider the problem, where $f(x) = \frac12 (x - \mu)^2$ that is 1-strongly convex, satisfies Assumption~\ref{as:L_smoothness}, and the oracle gives an unbiased gradient estimate.
    Also, we assume that the noise in the gradient estimate satisfies Assumption~\ref{as:bounded_bias_and_variance}.
    Then, there exists $C > 0$ such that the \algname{clipped-SGD} with learning rate $\gamma = \min \left( \frac{1}{400L \ln \frac{4(K+1)}{\delta}},  \frac{\ln ((K+1)R^2)}{K+1} \right)$ and clipping hyperparameter $\lambda_k = \frac{\exp(-\gamma (1 + k/2)) R}{120 \gamma \ln \frac{4(K+1)}{\delta}}$ provides the iterates such that after $k = 1,\ldots,K$ iterations the following bound holds with probability at least $1 - \delta$
    \[
    f(x_k) - f^* \leq  C \frac{\ln \frac{4(K+1)}{\delta} \ln^2 ((K+1)R^2)}{k+1}
    \]
    where $K$ is sufficiently large and $R \geq \|x_0 - x^*\|$.
\end{theorem}
The proof of this theorem is presented in Appendix~\ref{sec::proof-clip-sgd-th}.


\begin{corollary}\label{cor:SGD_biased_heavy_noise_str_cvx}
   Let Assumptions~\ref{as:bounded_bias_and_variance} and~\ref{as:L_smoothness}  hold on $Q = B_{2R}(x^*)$, where $R \geq \|x^0 - x^*\|$. 
   Suppose that $\nabla f_{\Xi^k}(x^k)$ satisfies Assumption~\ref{as:bounded_bias_and_variance}, and $\gamma = \min \left( \frac{1}{400L \ln \frac{4(K+1)}{\delta}},  \frac{\ln ((K+1)R^2)}{K+1} \right)$ 
    $
    \lambda_{k} \equiv \frac{\exp(-\gamma (1 + k/2)) R}{120 \gamma \ln \frac{4(K+1)}{\delta}}.
    $ 
    Then \algname{clipped-SGD} is $C \frac{\ln \frac{4(K+1)}{\delta} \ln^2 ((K+1)R^2)}{k+1}$-bounding first-order algorithm.
\end{corollary}

\subsubsection{\algname{Clipped-SGD-UCB}}

We are ready to present \algname{Clipped-SGD-UCB} algorithm.
For each arm $i$ ($i=1, \dots, K$) choose 
$$
f_i(x) = \frac{1}{2}(x-\mu_i)^2,
$$
with stochastic first-order oracle $\nabla f_i (x, \xi) = x-\xi$, where $\xi$ is a random variable sampled from $\mathcal{D}_i$. 

Then if Assumption~\ref{as:convolution} holds, the noise on gradient $\nabla f_i(x) - \nabla f_i(x, \xi) = \xi - \mu_i$ is symmetric.

\begin{algorithm}
\caption{ \algname{Clipped-SGD-UCB}}
\label{alg:clipped_sgd_ucb}
\begin{algorithmic}[1]
\REQUIRE{Number of arms $K$, period $T$, batch-size $b=(2m+1)n$, initial estimates $x_1^0=\dots=x_K^0=x^0$, clipping regime $\{\lambda_t\}_{t=1}^{\infty}$, learning rate schedule $\{\gamma_t\}_{t=1}^{\infty}$, parameter $\delta$}.
    \STATE Run \algname{Clipped-SGD-UCB} for each arm $i$ ($i=1, \dots, K$) independently for $b$ iterations and compute $\nabla_{\Xi_i^0} f_i(x_i^0)$ and $x_i^1 = x_i^0 - \gamma_0 \clip(\nabla f_{\Xi^1_i}(x_i^0), \lambda_0)$.
    \STATE For each arm $i$ ($i=1, \dots, K$) set $n_i(K)=1$ and compute $UCB(i, n_i(K), \delta) = x_i^1 + \sqrt{g(1, \delta)}$.
    \FOR{$t=1, \dots, T$}
    \STATE Choose arm $i_t = \arg \max_{1 \leq i \leq K} UCB(i, n_i, \delta)$.
    \STATE Play $i_t$ arm $b$ times, observe rewards and compute $\nabla_{\Xi_{i_t}^{n_{i_t}}} f_{i_t}(x_{i_t}^{n_{i_t}})$.
    \STATE Compute $x_{i_t}^{n_{i_t}+1} = x_{i_t}^{n_{i_t}} - \gamma_{n_{i_t}} \clip(\nabla_{\Xi_{i_t}^{n_{i_t}}} f_i(x_{i_t}^{n_{i_t}}), \lambda_{n_{i_t}})$.
    \STATE Set $n_{i_t}(t+1) = n_{i_t}(t)+1$ (increase counter by one).
    \STATE Set 
    $$
    UCB(i, n_i(t+1), \delta) = \begin{cases}
    &UCB(i, n_i(t), \delta), \quad i\neq i_t,\\
    &x_{i_t}^{n_{i_t}} + \sqrt{2g(n_{i_t}(t+1), \delta)}, \quad \text{otherwise}.
    \end{cases}
    $$
    \ENDFOR
\end{algorithmic}
\end{algorithm}

\begin{theorem}[Convergence of \algname{Clipped-SGD-UCB}]\label{thm:MAB_theorem} 
 The regret of the \algname{Clipped-SGD-UCB} for multi-armed bandit problem with $K$ arms, period $T$, $\gamma = \min \left( \frac{1}{400L \ln \frac{4(K+1)}{\delta}},  \frac{\ln ((K+1)R^2)}{K+1} \right)$, $\lambda_k= \frac{\exp(-\gamma (1 + k/2)) R}{120 \gamma \ln \frac{4(K+1)}{\delta}}$ and symmetric distribution of rewards satisfies:
  $$
R_T \leq 4\log((T+1)TR^2)\sqrt{C \log(4T(T+1)^2)TK} +  \sum_i \Delta_i $$
$$
R_T \leq \sum_{i: \Delta_i>0}  \left [ \Delta_i + \frac{8C \log(4T(T+1^2)) \log^2((T+1)TR^2)}{\Delta_i}\right ] 
$$

\end{theorem}
The proof of this theorem is presented in Appendix~\ref{sec::proof-MAB-theorem}.








\begin{remark} 
If the algorithm uses a batch of samples to perform a single step  with the batch size $b$, the regret will increase in $b$ times, but the number of iterations will be $\frac{T}{b}$. 
Thus adaptive part of the bound will not change, and only the fixed part will increase, i.e. each arm will require at least $2b$ samples instead of $2$.
\end{remark}


\section{Numerical Experiments}
\label{sec:NumExp}

In this section, we demonstrate the superior performance of the proposed algorithm in the following environments.
The main feature of the test environment is the structure of the noise that models the uncertainty of the observed rewards.
In the simulations of the multi-armed bandit, one can obtain a reward estimate $r_i = \mu_i + \xi$ corresponding to the $i$-th arm, where $\mu_i$ is the ground-truth reward and $\xi$ is the aforementioned noise, whose distribution is the key feature of the testing environments.  
In particular, we focus on super-heavy and heavy tail noise distributions.
Also, we test Gaussian noise to show the performance of our algorithm in a simple environment.
The additional feature of the environment is the number of arms and the distribution of the corresponding ground-truth rewards.
The closer these rewards are, the more challenging the MAB problem is. 
For a better illustration of the algorithms' performance, we adjust the particular instances of such environments for the considered noise structure and provide details in the corresponding sections.

We compare the \texttt{UCB} algorithm~\cite{auer2002finite}, the Robust UCB algorithm~\cite{bubeck2013bandits} that uses median of means to estimate rewards (we further refer to this algorithm as \texttt{RUCB-Median}), \texttt{SGD-UCB}, \texttt{SGD-UCB-Median} and \texttt{SGD-UCB-SMoM} algorithms.
The latter three algorithms are the particular instances of the proposed framework summarized in Algorithm~\ref{alg:clipped_sgd_ucb}.
In particular, \texttt{SGD-UCB} corresponds to the values $m=0, n=1$, uses the single sample to estimate gradient similar to vanilla SGD. 
\texttt{SGD-UCB-Median} corresponds to the values $m=1, n=1$, takes three samples and uses their median to estimate gradient. 
\texttt{SGD-UCB-SMoM} corresponds to the values $m=1, n=2$ and uses SMoM~\eqref{eq::smom_def} as a gradient estimate.
The parameters $m$ and $n$ are used in Algorithm~\ref{alg:clipped_sgd_ucb} to generate batch size $b$ and construct gradient estimate according to Definition~\ref{def:smom}.
We do not consider the Truncated Robust UCB algorithm~\cite{bubeck2013bandits} since it shows worse performance compared to the \texttt{RUCB-Median} algorithm.
The source code for reproducing the presented results can be found in the GitHub repository~\url{https://github.com/tmpuser1233/Clipped-SGD-UCB}.




\paragraph{Initialization of reward estimates.}
To initialize the reward estimate for every arm we use the following procedure.
Every arm is pulled $p$ times ($p$ is an odd number) and the median of the obtained rewards are used as initialization $x_i^1$ in the notation of Algorithm~\ref{alg:clipped_sgd_ucb}.
In this setup we skip the line 1 in pseudocode presented in Algorithm~\ref{alg:clipped_sgd_ucb}.
From our experience we recommend using $p = 1$ for Gaussian rewards noise and $p=3$ for heavy and super-heavy tailed rewards nose.




\subsection{Super-heavy tail MAB}
In this section, we consider the super-heavy tail distributions of the noise used in the rewards uncertainty simulation.
A distribution has a super-heavy tail if the expectation of the corresponding random variable does not exist.
In particular, we test Cauchy distributions with the CDF $p_C(x) = \frac{1}{\pi\gamma[1 + (\frac{x}{\gamma})^2]}$, where $\gamma = 1$, Fr\'echet distribution with the CDF $p_F(x) = e^{-x^{-\alpha}}$, where $\alpha = 1$, the mixture of Cauchy ($\gamma = 1$) and exponential distributions with the CDF $p_{CE}(x) = 0.7 \cdot p_C(x) + 0.3 \cdot e^{-(x + 1)}\mathbb{I}\{x \ge -1\}$ and the mixture of Cauchy ($\gamma = 1$) and Pareto distributions with the CDF $p_{CP} = 0.7 \cdot p_C(x) + 0.3 \cdot \frac{3}{(x + 1.5)^4}\mathbb{I}\{x \ge -1.5\}$.
Note that the latter two mixtures of distributions represent the asymmetric distributions.
Although we do not consider asymmetric noise in our proof, we demonstrate the performance of the proposed framework for such noise distributions empirically.

To simulate multi-armed bandit, we use the following three environments: 10 arms and the ground-truth reward of the $i$-th arm $\mu_i = i, i = 0, \ldots, 9$, 10 arms and the ground-truth reward of the $i$-th arm $\mu_i = i/10, i = 0, \ldots, 9$, and 100 arms and the ground-truth reward of the $i$-th arm $\mu_i = i/50, i = 0, \ldots, 99$.
Further, we refer to these environments as \texttt{Env1}, \texttt{Env2} and \texttt{Env3}, respectively. 

\paragraph{Convergence comparison.}

\begin{figure}[!h]
    \centering
    \begin{subfigure}{0.3\linewidth}
\includegraphics[width=\linewidth]{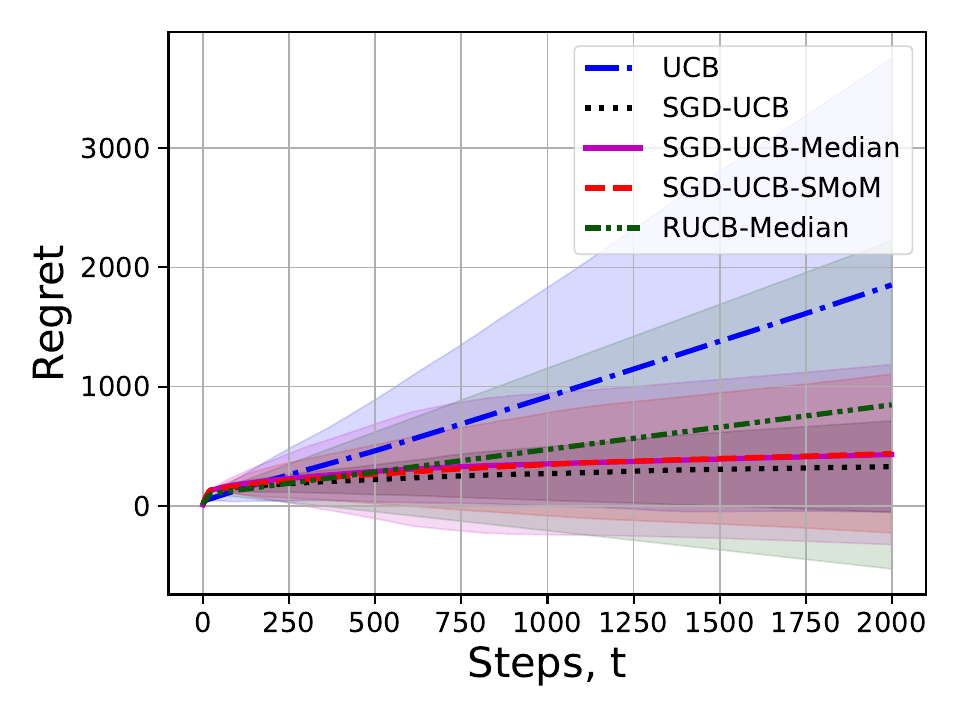}
\caption{\texttt{Env1}}
    \end{subfigure}
    ~
    \begin{subfigure}{0.3\linewidth}
\includegraphics[width=\linewidth]{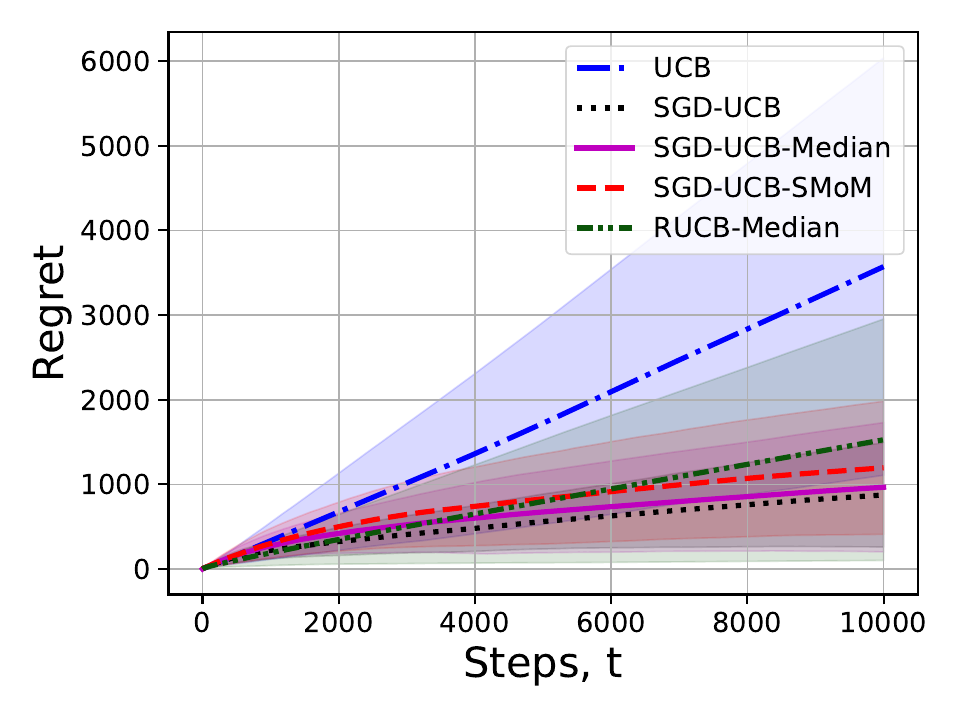}
\caption{\texttt{Env2}}
    \end{subfigure}
    ~
    \begin{subfigure}{0.3\linewidth}
\includegraphics[width=\linewidth]{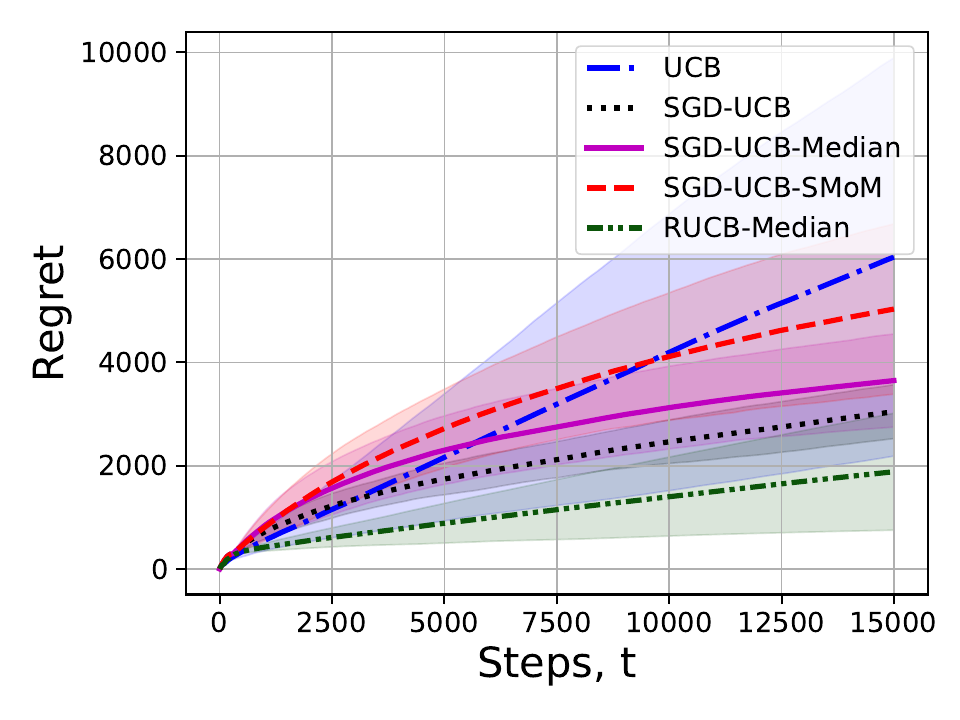}
\caption{\texttt{Env3}}
    \end{subfigure}
\\
\begin{subfigure}{0.3\linewidth}
\includegraphics[width=\linewidth]{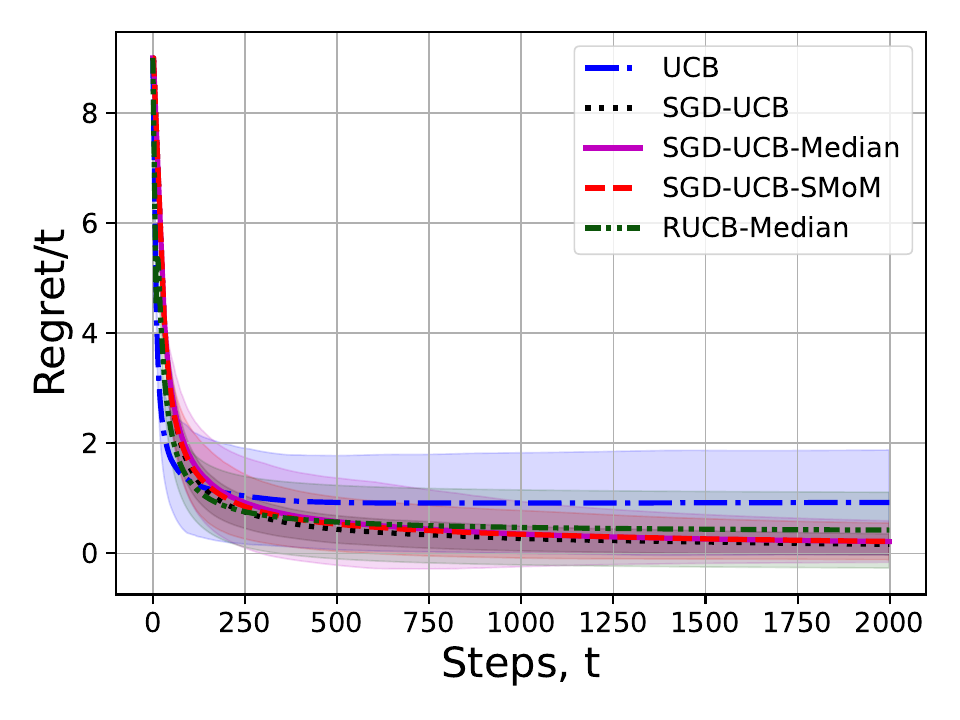}
\caption{\texttt{Env1}}
\end{subfigure}
~ 
\begin{subfigure}{0.3\linewidth}
\includegraphics[width=\linewidth]{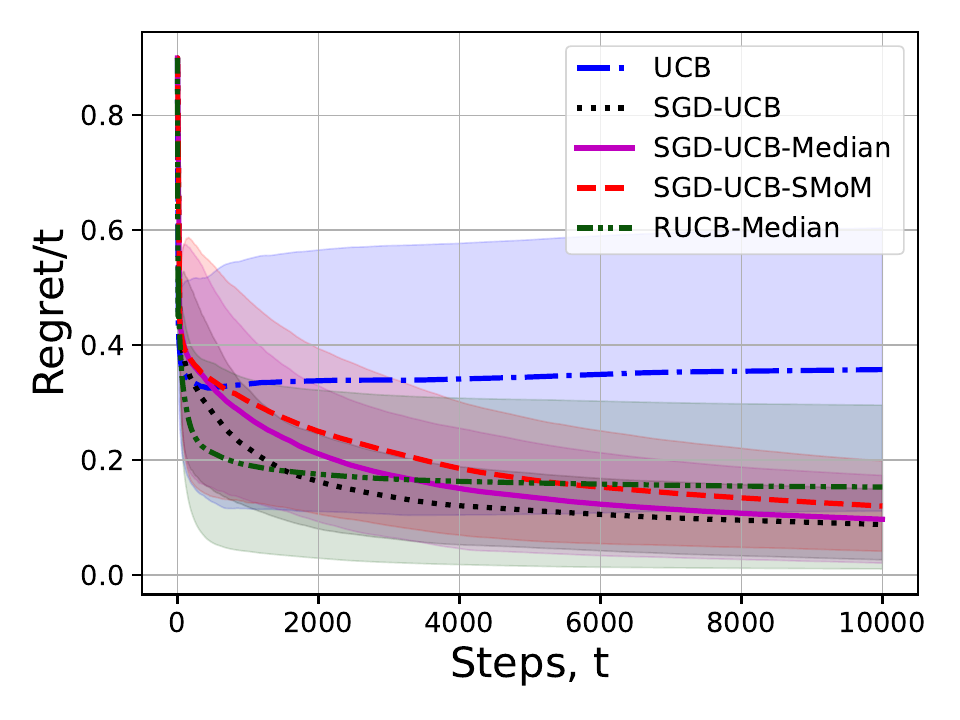}
\caption{\texttt{Env2}}
\end{subfigure}
~
\begin{subfigure}{0.3\linewidth}
\includegraphics[width=\linewidth]{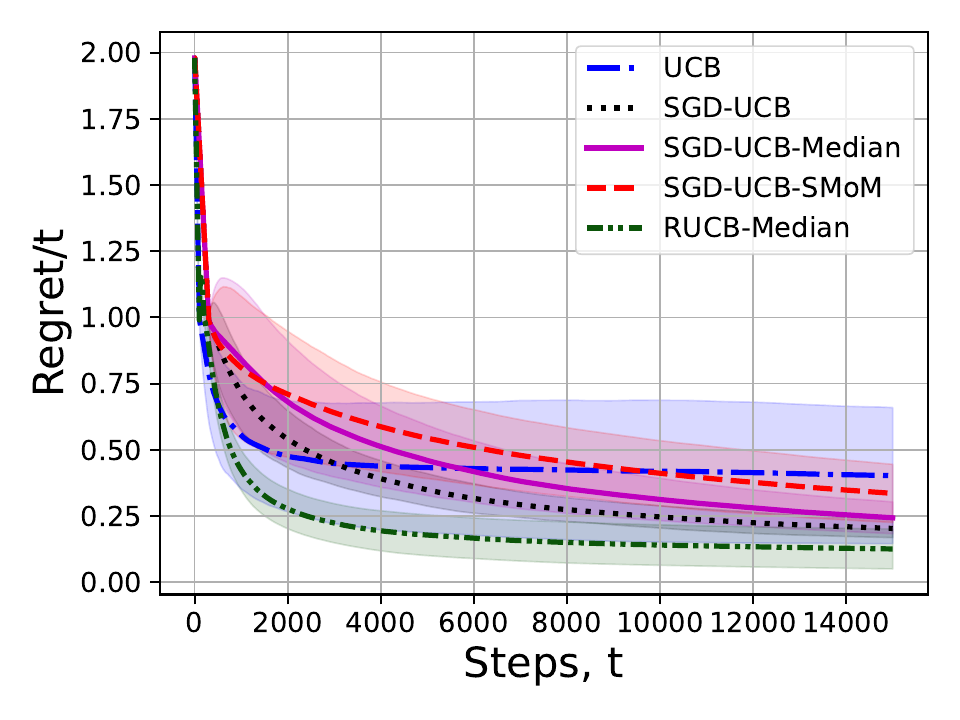}
\caption{\texttt{Env3}}
\end{subfigure}
    \caption{The convergence of the regret metric (the first row) and the mean regret metric (the second row) for the considered algorithms with Cauchy distribution ($\gamma=1$) of a reward noise. We report the averaged values over 120 trials and the corresponding standard deviation area via shaded regions. Our algorithms show faster convergence in \texttt{Env1} and \texttt{Env2} compared to competitors and slightly slower convergence than \texttt{RUCB-Median} in \texttt{Env3}.}
    \label{fig::super-heavy-tail}
\end{figure}

To compare the convergence of the considered algorithms we test three environments mentioned above. 
Due to the space limitation, we provide here only plots corresponding to the Cauchy distribution ($\gamma = 1$) of the reward noise $\xi$.
The similar plots corresponding to the super-heavy tail distributions are presented in Appendix~\ref{sec::app_exp_heavy}.
We use the hyperparameters of the algorithms which give the best convergence.
Figure~\ref{fig::super-heavy-tail} shows that the proposed algorithms outperform \texttt{RUCB-Median} and \texttt{UCB} algorithms in \texttt{Env1} and \texttt{Env2}.
At the same time, \texttt{Env3} appears more challenging, and \texttt{RUCB-Median} shows slightly faster convergence in terms of the number of steps.
Despite this, we show in the next paragraph (see Table~\ref{tab:runtime}) that our algorithms  are significantly faster in terms of runtime since the single iteration costs are significantly smaller.


%

\paragraph{Runtime comparison.}
In addition to the convergence comparison presented in Figure~\ref{fig::super-heavy-tail}, we also provide the runtime comparison of the considered algorithms.
Such comparison is important for highlighting the difference in the costs for a single step in the discussed algorithms. The design of this experiment is the following. 
We assign to every algorithm the budget for total pulls of arms equal to $10^4$. 
In simulations, we track the mean regret $R_T / T$ and measure the runtime to achieve the target values of this metric. 
We test the target metrics $R_T/T = 0.1$ and $R_T/T = 0.05$. 
If an algorithm does not achieve the target value of the mean regret within the assigned budget, we consider such a run as a fail.
We run 100 trials for every algorithm and show in Table~\ref{tab:runtime} the 90\% percentile of their running time.

\begin{table}[!h]
    \centering
    \caption{Runtime comparison of the considered algorithms to achieve the listed values of $R_T/T$ for the environment \texttt{Env1} with Cauchy distribution ($\gamma=1$) as a reward noise. Our algorithms more often reach the target mean regret within the assigned budget, i.e. the number of failed trials is smaller. Also, we show that our algorithms are significantly faster than \texttt{RUCB-Median}. We highlight the best values for runtime and \# fails with bold.}
    \scalebox{0.9}{
    \begin{tabular}{ccccc}
    \toprule
      Algorithms & Runtime for $R_T/T = 0.1$, s. & \# fails & Runtime for $R_T/T = 0.05$, s & \# fails\\
    \cmidrule(lr){1-1} \cmidrule(lr){2-3} \cmidrule(lr){4-5}
      \texttt{SGD-UCB}   &   2.8  & \textbf{7} & 3.7 & \textbf{17}\\
      \texttt{SGD-UCB-Median} & 2.4 & 17 & 2.6 & 18\\
      \texttt{SGD-UCB-SMoM} & \textbf{1.3} & 12 &  \textbf{1.4} & 30 \\
      \texttt{RUCB-Median} & 32.9 & 37 &  33.3 & 37\\
    \bottomrule
    \end{tabular}
    }
    \label{tab:runtime}
\end{table}

Note that although the numbers presented in Table~\ref{tab:runtime} depend on environments, the complexity of the single step of every algorithm preserves over the different environments.
Therefore, we can expect that the ordering of the algorithms in terms of the runtime will be the same for other noise and environments.







\subsection{Heavy-tail MAB}

To test the proposed framework in the heavy-tail MAB problem setup, we use the similar environments as in the previous section and Fr\'echet distribution with the CDF $p_F(x) = e^{-x^{-\alpha}}$, where $\alpha = 1.25$ to model the noise in the reward estimates.

\begin{figure}[!h]
    \centering
    \begin{subfigure}{0.3\linewidth}
\includegraphics[width=\linewidth]{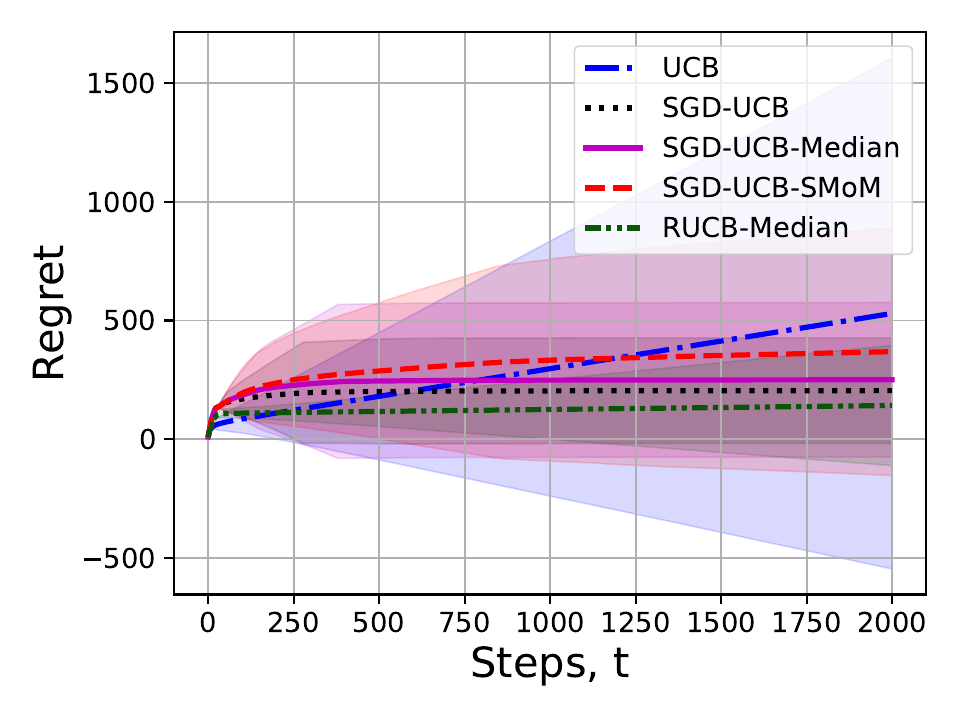}
\caption{\texttt{Env1}}
    \end{subfigure}
    ~
    \begin{subfigure}{0.3\linewidth}
    \includegraphics[width=\linewidth]{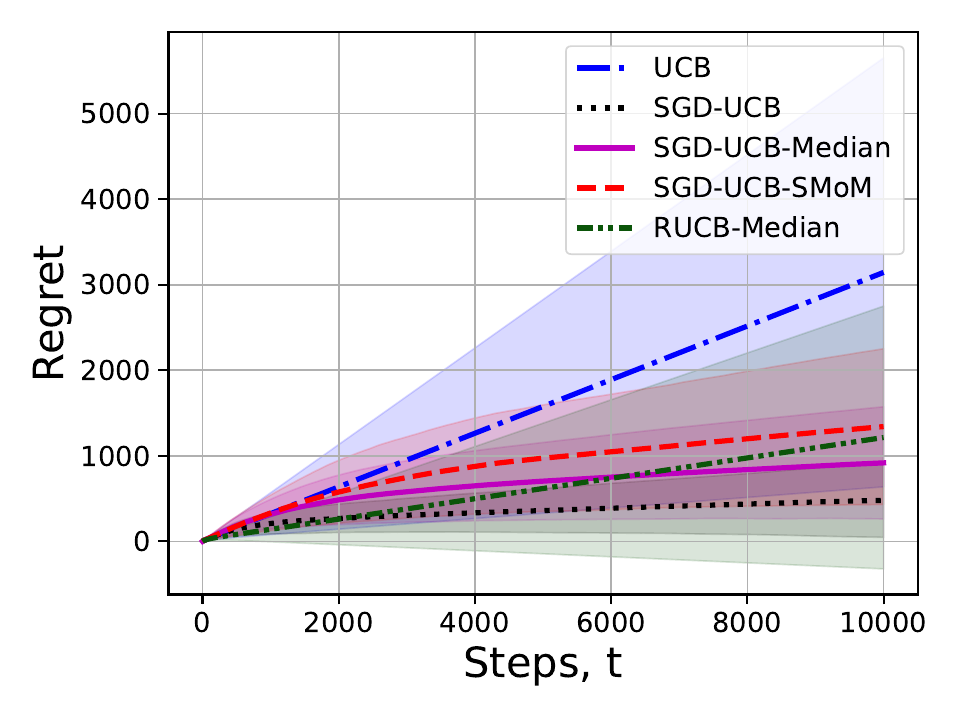}
\caption{\texttt{Env2}}
    \end{subfigure}
    ~
    \begin{subfigure}{0.3\linewidth}
    \includegraphics[width=\linewidth]{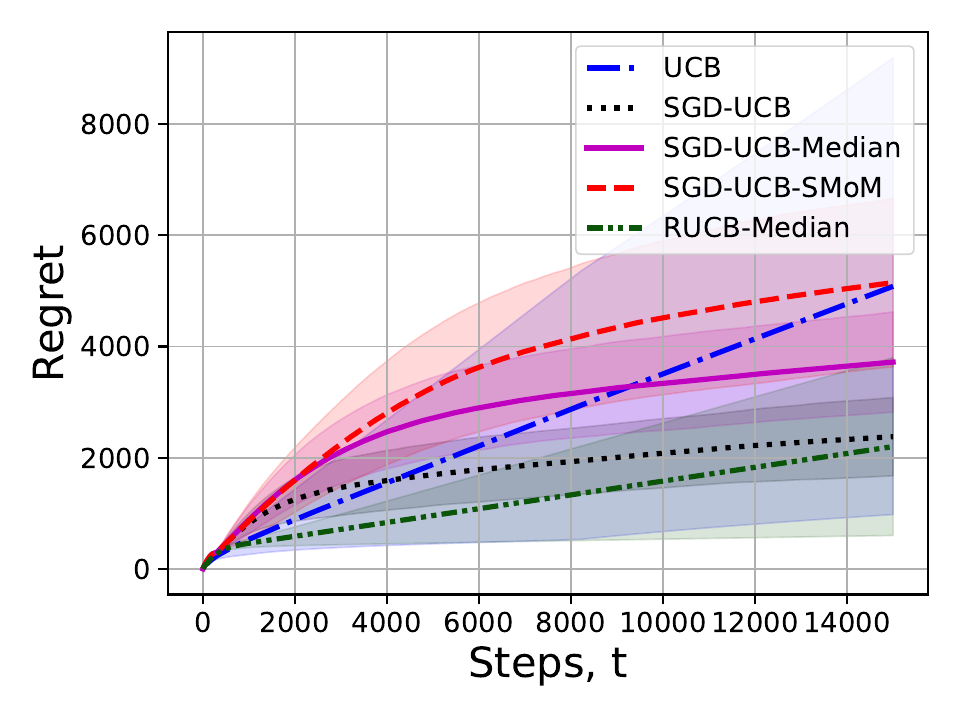}
\caption{\texttt{Env3}}
    \end{subfigure}
    \\
    \begin{subfigure}{0.3\linewidth}
    \includegraphics[width=\linewidth]{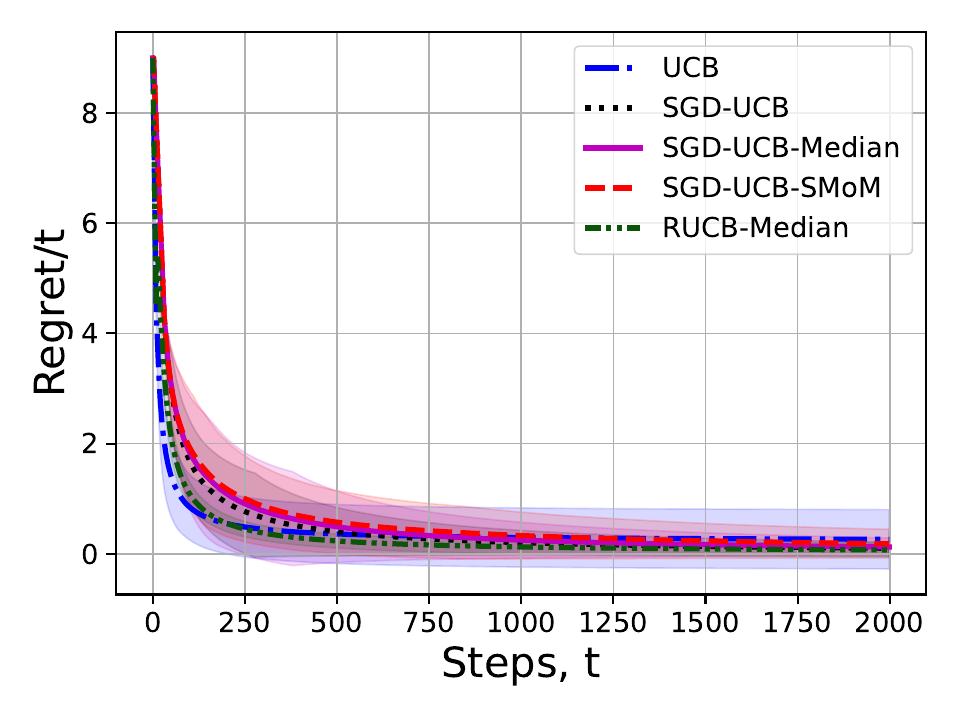}
\caption{\texttt{Env1}}
    \end{subfigure}
    ~
    \begin{subfigure}{0.3\linewidth}
    \includegraphics[width=\linewidth]{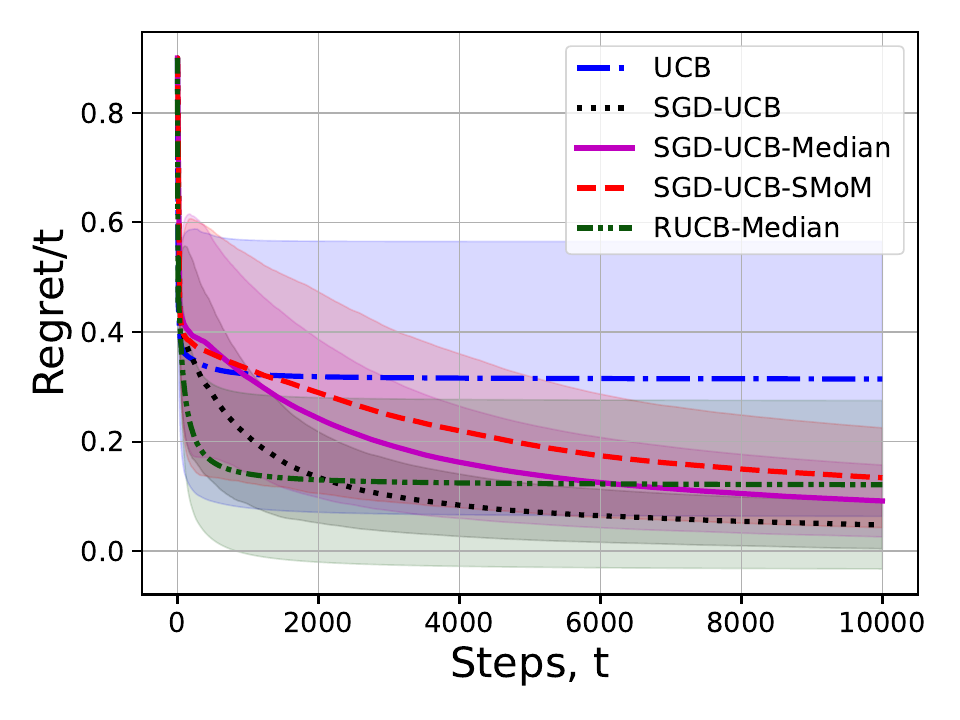}
\caption{\texttt{Env2}}
    \end{subfigure}
    ~
    \begin{subfigure}{0.3\linewidth}
    \includegraphics[width=\linewidth]{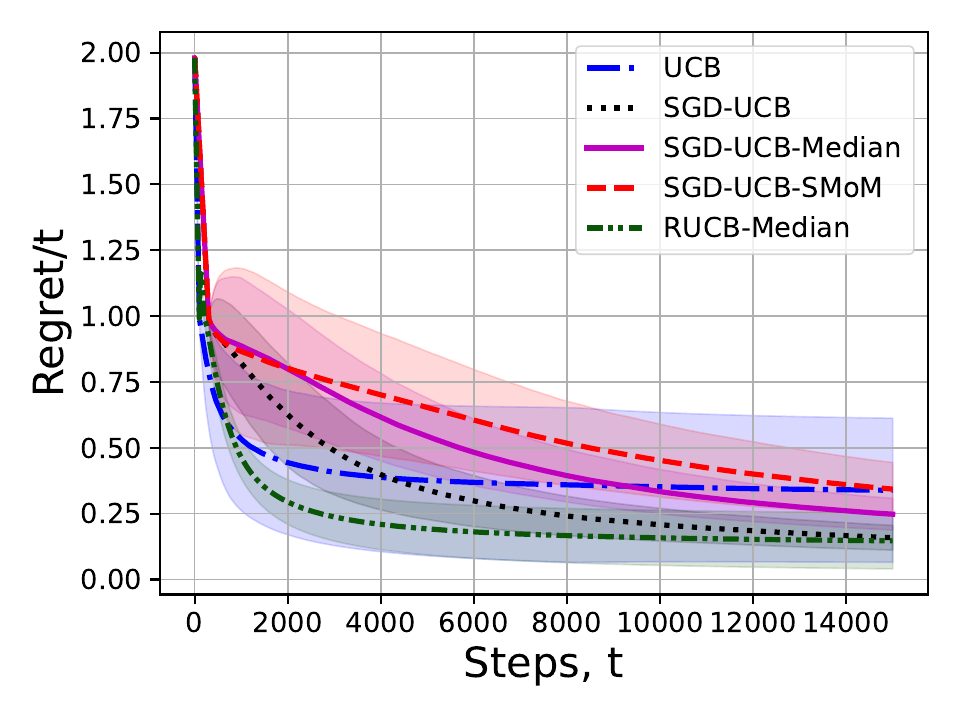}
\caption{\texttt{Env3}}
    \end{subfigure}
    \caption{The convergence of the regret metric (the first row) and the mean regret metric (the second row) for the considered algorithms with Fr\'echet distribution ($\alpha=1.25$) of a reward noise. We report the averaged values over 120 trials and the corresponding standard deviation area via shaded regions. Our algorithms show faster convergence in \texttt{Env2}, asymptotically faster convergence in \texttt{Env3}, and slower convergence than \texttt{RUCB-Median} in \texttt{Env1}.}
    \label{fig::heavy-tail-mab}
\end{figure}

Figure~\ref{fig::heavy-tail-mab} shows that our algorithms (\texttt{SGD-UCB} and \texttt{SGD-UCB-Median}) provide smaller mean regret for the considered number of steps in \texttt{Env2} than \texttt{RUCB-Median} and \texttt{UCB} algorithms.
At the same time, in \texttt{Env3} we observe only the asymptotically faster convergence of \texttt{SGD-UCB} and \texttt{SGD-UCB-Median} compared to the \texttt{RUCB-Median} algorithm.
In the observed number of steps, the \texttt{RUCB-Median} algorithm provides smaller values of mean regret.
In addition, \texttt{Env1} is especially challenging for the proposed algorithms. 
The \texttt{RUCB-Median} algorithm outperforms them and provides the smaller regret values in the considered number of steps.
However, the difference between the regret given by the \texttt{RUCB-Median} algorithm and the \texttt{SGD-UCB} algorithm is not large and the corresponding mean regret values are already almost the same.

\subsection{Gaussian MAB}
\label{sec::gauss_mab}



\begin{figure}[!h]
    \centering
    \begin{subfigure}{0.3\linewidth}
        \includegraphics[width=\linewidth]{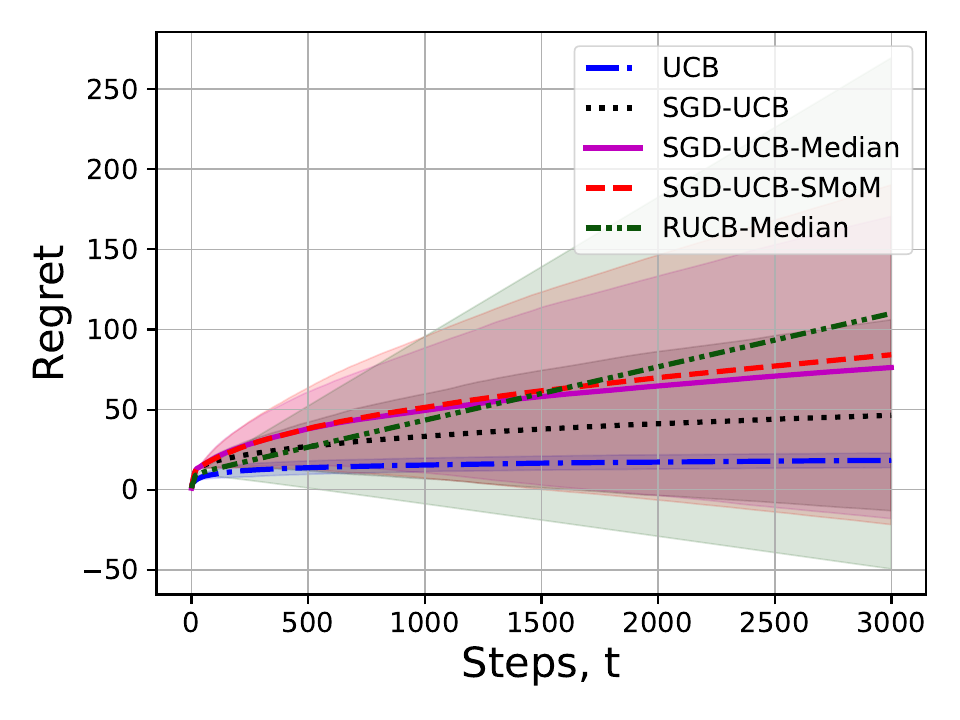}
    \end{subfigure}
    ~
    \begin{subfigure}{0.3\linewidth}
        \includegraphics[width=\linewidth]{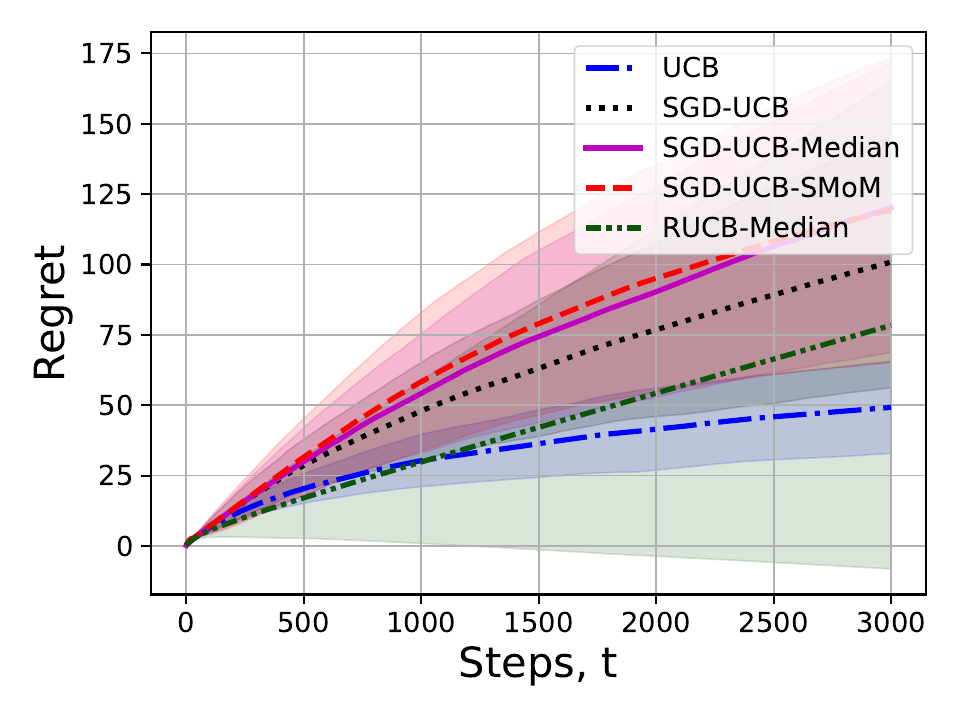}
    \end{subfigure}
    ~
    \begin{subfigure}{0.3\linewidth}
        \includegraphics[width=\linewidth]{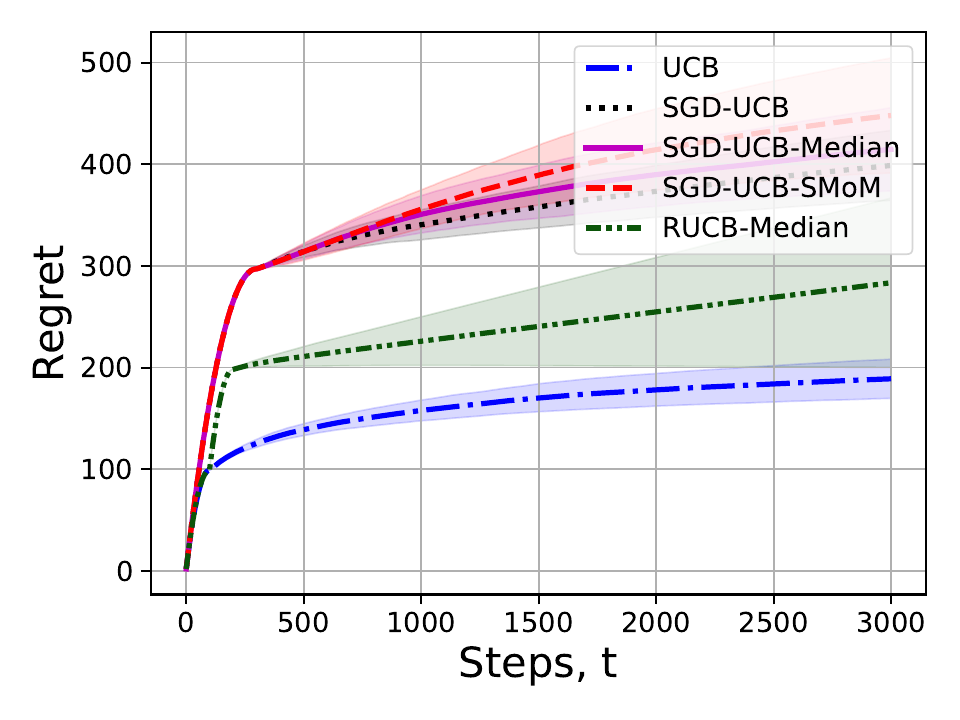}
    \end{subfigure}
    \\
    \begin{subfigure}{0.3\linewidth}
        \includegraphics[width=\linewidth]{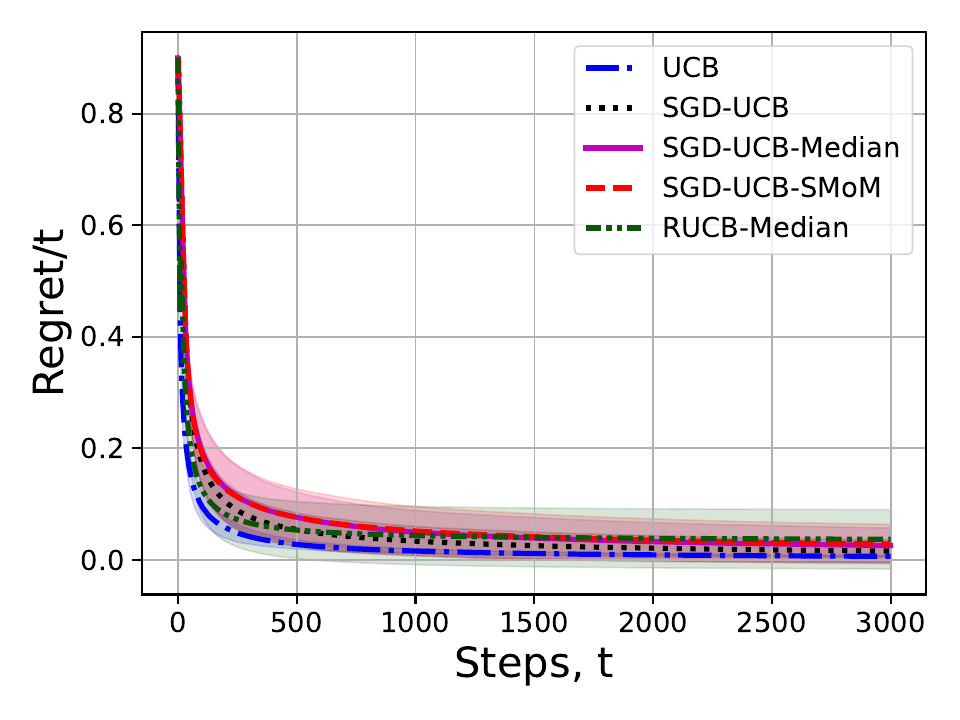}
    \end{subfigure}
    ~
    \begin{subfigure}{0.3\linewidth}
        \includegraphics[width=\linewidth]{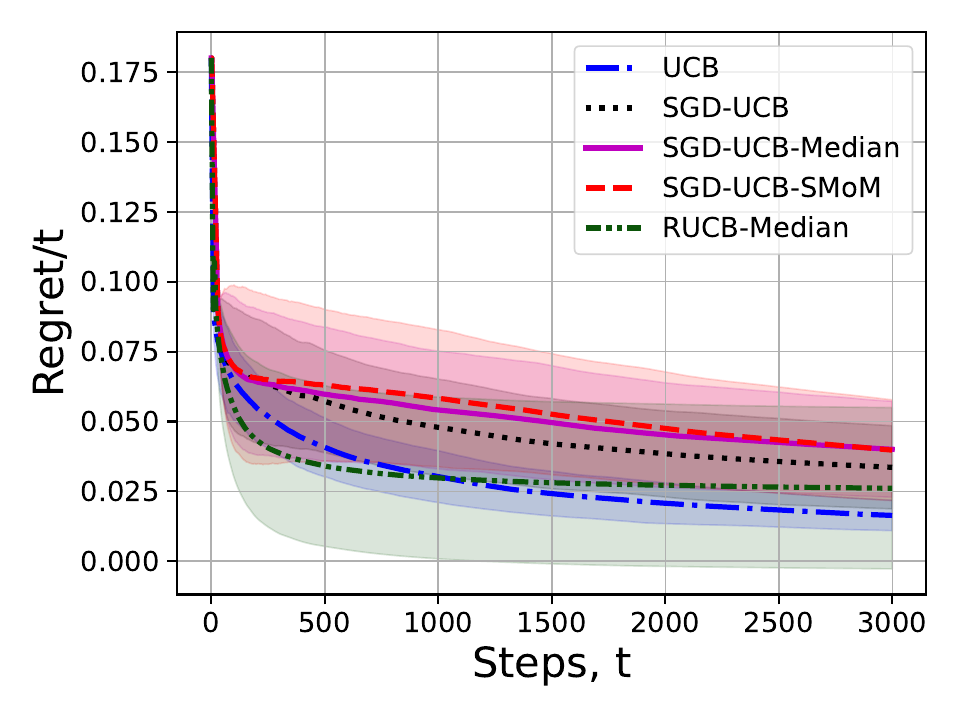}
    \end{subfigure}
    ~
    \begin{subfigure}{0.3\linewidth}
        \includegraphics[width=\linewidth]{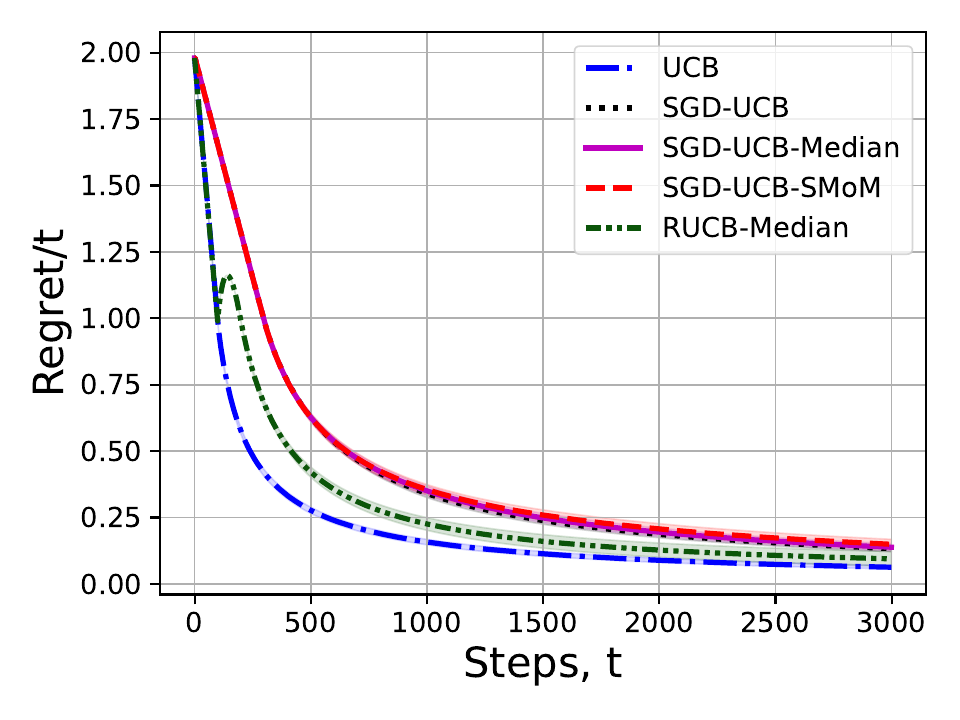}
    \end{subfigure}
    \caption{Convergence of the considered algorithms in test environments: a) 10 arms and $\mu_i = i / 10$, where $i=0, \ldots, 9$ (the first column); b) 10 arms and $\mu_i = i / 50$, where $i=0, \ldots, 9$ (the second column); c) 100 arms and $\mu_i = i / 50$, where $i=0, \ldots, 99$ (the third column). In such test environments, \texttt{UCB} provides the best regret and the mean regret compared to alternatives while our algorithms converge to the same limit values of the mean regret.}
    \label{fig::standard_gaussian}
\end{figure}

In this section, we consider three simple environments to test multi-armed bandits, where reward noise is generated from the standard normal distribution $\mathcal{N}(0, 1)$.
The first one corresponds to a bandit with 10 arms and the reward for the $i$-th arm is computed as follows $\mu_i = i / 10$, where $i=0, \ldots, 9$. 
The second one corresponds to a bandit with 10 arms and the reward for the $i$-th arm is computed as follows $\mu_i = i / 50$, where $i=0, \ldots, 9$. 
The third one corresponds to a bandit with 100 arms and the reward for the $i$-th arm is computed as follows $\mu_i = i / 50$, where $i=0, \ldots, 99$.
We run 150 trial simulations for 3000 steps and average the final regret.
The results of the comparison are presented in Figure~\ref{fig::standard_gaussian}.
These plots show that in the case of the Gaussian reward noise, the smallest regret is given by the vanilla \texttt{UCB} algorithm uniformly for the considered environments.
However, the proposed algorithms still converge almost to the same mean regret values, where \texttt{UCB} converges.
This experiment demonstrates that the proposed algorithms show similar performance to the optimal algorithm in the Gaussian reward noise setup.


\subsection{MAB problem with hardly distinguished arms}

\begin{figure}[!h]
    \centering
    \begin{subfigure}{0.4\textwidth}
        \includegraphics[width=\linewidth]{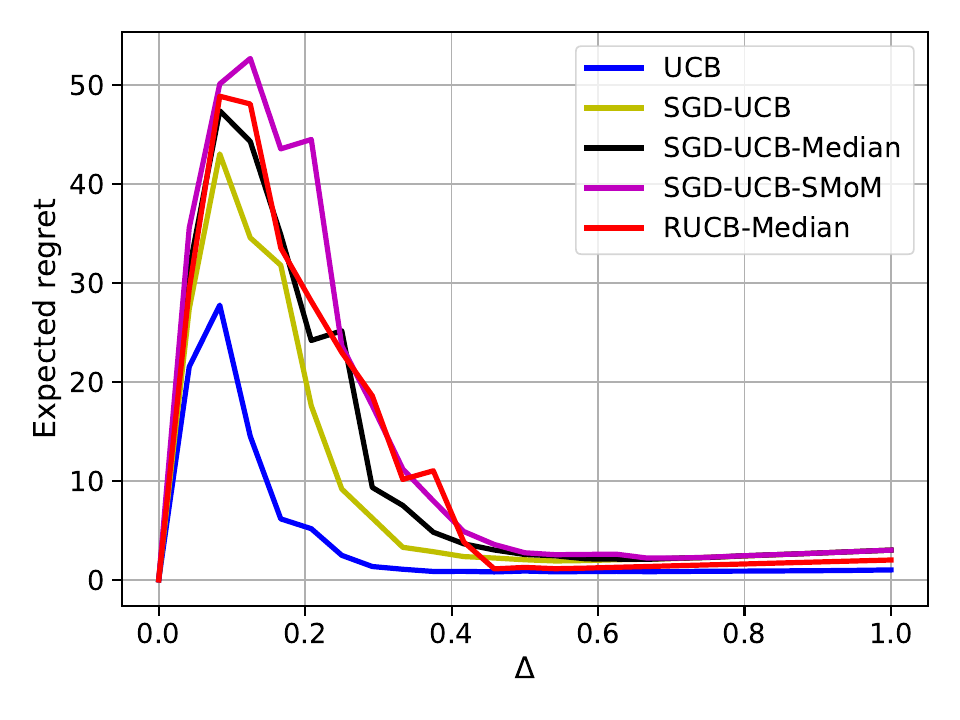}
        \caption{Gaussian MAB}
    \end{subfigure}
    ~
    \begin{subfigure}{0.4\textwidth}
        \includegraphics[width=\linewidth]{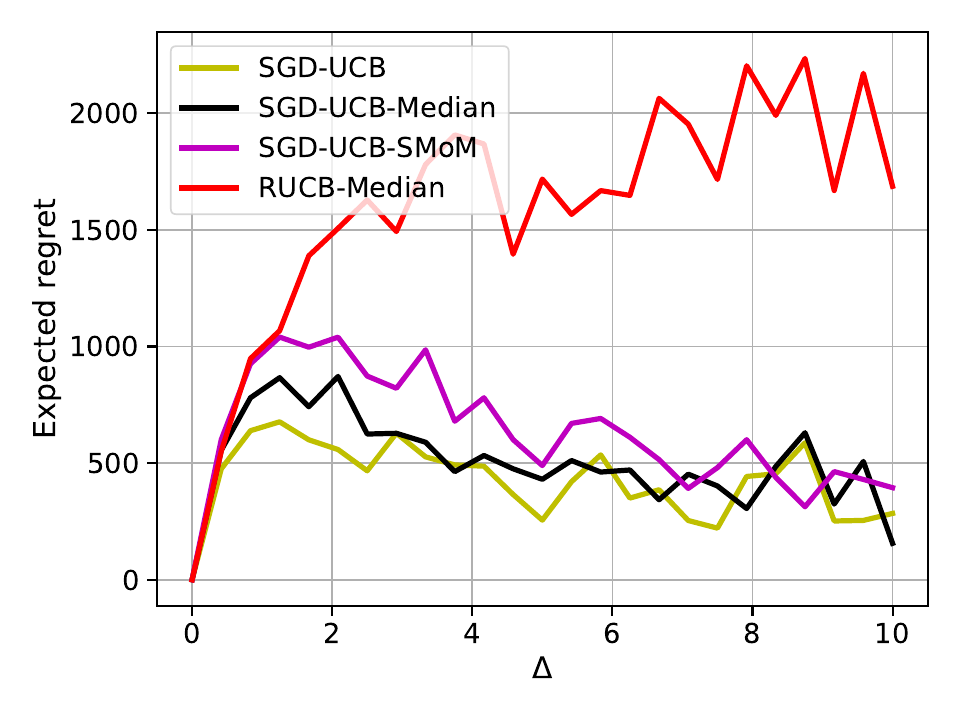}
        \caption{Heavy tail MAB}
    \end{subfigure}
    \caption{(a) Comparison of the mean regret for the considered algorithms in Gaussian MAB with two arms with rewards $\{0,\Delta\}$. 
    Our algorithms can distinguish arms with close rewards similar to the competitors. (b) Comparison of the mean regret for the considered algorithms in heavy tail MAB with five arms with rewards $\{0,0,0,0,\Delta\}.$ Our algorithms can distinguish arms with close rewards even if the noise is generated from the Cauchy  distribution ($\gamma = 1$).}
    \label{fig::delta_light}
\end{figure}

In addition, to evaluate the robustness of the considered algorithms to the closed ground-truth rewards, we consider the bandit with hardly distinguishable rewards. 
For Gaussian MAB we consider two arms such that the corresponding rewards are $\{0, \Delta\}$. 
The values of $\Delta$ vary from 0 to 1 with a step size of 0.04. 
For heavy tail MAB we consider Cauchy distribution ($\gamma = 1$) with 5 arms such that the corresponding rewards are $\{0, 0,0,0,\Delta\}$. 
The values of $\Delta$ vary from 0 to 10 with a step size of 0.4.
For each environment setup, we run 300 trial simulations for 2000 steps and average the final regret on trials.
The result of the robustness analysis is presented in Figure~\ref{fig::delta_light}. 
It shows that UCB is optimal for Gaussian MAB, and our algorithms are close to the \texttt{RUCB-Median} algorithm in terms of the expected regret for close arms rewards.
In the heavy tail environment, we do not plot UCB, since its expected regret grows crucially and suffers readability. 
Our algorithms show smaller expected regrets as arms become more distinguishable. 

\section{Conclusion and future work}

We suggested a new template to construct UCB-type algorithms for stochastic multi-armed bandits with heavy tails. 
The main ingredient is to use $g(k, \delta)$-bounded algorithms for optimization problems with inexact oracle. 
As the main illustration example, we propose \algname{Clipped-SGD-UCB} algorithm. 
The proposed algorithm shows convergence even in the case of noise, which has no expectation.
Future work includes the construction of more appropriate algorithms fine-tuned for the proposed template. 
It is also interesting to find good and nontrivial zero-order algorithms appropriate to the proposed template.


\bibliographystyle{unsrt}
\bibliography{lib}

\appendix

\section{Proof of Theorem~\ref{thm:FO_UCB}}
\label{sec::proof-FO-UCB}
Here we provide the detailed proof of Theorem~\ref{thm:FO_UCB} and its formulation for the reader's convenience.
\setcounter{theorem}{1}
\begin{theorem}[Convergence of \algname{FO-UCB}]
 The regret of the \algname{FO-UCB} with $g(k, \delta)$-bounded first-order algorithm for the MAB problem with $K$ arms, auxiliary functions $f_i(x) = \frac{1}{2}(x-\mu_i)^2$, period $T$, $\delta=\frac{1}{T (T+1)}$ satisfies
\begin{align}
    &R_T \leq \sum_{i: \Delta_i>0} \Delta_i \left (g^{-1}\left (\frac{\Delta_i^2}{8}, \frac{1}{T^2} \right ) + 1 \right )
\end{align}
\end{theorem}

\proof

For proof, we follow the standard approach (see \cite{lattimore2020bandit}).
Denote by $\Delta_i = \mu_{i^*} - \mu_i$, where $i^* = \arg \max_{1\leq i \leq K} \mu_i$. Then regret can be computed as
$$
R_T = \sum_{i=1}^K \Delta_i \EE[n_i(T)]
$$
where $n_i(t)$ is the number of rounds before round $t$ when arm $i$ was chosen.

Let $G_i$ be a "good" event defined by
$$
G_i = \left \{\mu_{i^*} < \min_{1\leq t \leq T} UCB (i^*, n_{i^*}(t), \delta) \right \} \cap \left \{ UCB (i, u_i, \delta) < \mu_{i^*} \right \},
$$
where the constant $u_i$ will be chosen later.

We show that if $G_i$ holds, then $n_i(T) \leq u_i$. 
We assume that this is not true and $n_i(T) > u_i$. 
Then, there exists a round $t \leq T$ such that $n_i(t-1) = u_i$ and $A_t = i$. 
Then 
\begin{align*}
    &UCB(i, n_i(t-1), \delta) = x_i^{u_i} + \sqrt{2g(u_i, \delta)}<\mu_{i^*} < UCB(i^*, n_{i^*}(t-1), \delta).
\end{align*}
Hence, $A_t = \arg \max_{1 \leq j \leq K} UCB(j, n_j(t-1), \delta) \neq i$ and we obtain a contradiction.

Next, we bound the probability of the complement event 
$$
\hat{G}_i = \left \{\mu_{i^*} > \min_{1 \leq t \leq T} UCB(i^*, n_{i^*}(t), \delta)\right \} \cup \left \{x_i^{u_i} + \sqrt{2g(u_i, \delta)}>\mu_{i^*} \right \}.
$$
We can then determine the probability of the first term using a union bound:
\begin{align*}
    &\mathbb{P} \left [\mu_{i^*} > \min_{1 \leq t \leq T} UCB(i^*, n_{i^*}(t), \delta)\right ] \\
    &=\mathbb{P} \left [ \cup_{s \leq T} \left \{\mu_{i^*} > UCB (i^*, n_{i^*}(s), \delta) \right \} \right ]\\
    &\leq \sum_{s\leq T}\mathbb{P} \left [ \mu_{i^*} > UCB (i^*, n_{i^*}(s), \delta) \right ] \leq \delta T
\end{align*}
To bound the probability of the second term, we use the following scheme:
\begin{align*}
    &\mathbb{P} \left [x_i^{u_i} + \sqrt{2g(u_i, \delta)}>\mu_{i^*} \right ]\\
    &=\mathbb{P} \left [x_i^{u_i} - \mu_i + \sqrt{2g(u_i, \delta)}>\mu_{i^*} - \mu_i \right ] \\
    &=\mathbb{P} \left [x_i^{u_i} - \mu_i > \Delta_i - \sqrt{2g(u_i, \delta)} \right ]\\
    &=\mathbb{P} \left [x_i^{u_i} - \mu_i > \Delta_i - \sqrt{2g(u_i, \delta)} | \quad |x_i^{u_i} - \mu_i |> \sqrt{2g(u_i, \delta)} \right ] \cdot \mathbb{P} \left [ |x_i^{u_i} - \mu_i| > \sqrt{2g(u_i, \delta)} \right ]\\
    &+\mathbb{P} \left [x_i^{u_i} - \mu_i > \Delta_i - \sqrt{2g(u_i, \delta)} | \quad |x_i^{u_i} - \mu_i| \leq  \sqrt{2g(u_i, \delta)} \right ] \cdot \mathbb{P} \left [ |x_i^{u_i} - \mu_i |\leq \sqrt{2g(u_i, \delta)} \right ]\\
    &(\text{choose $u_i$: $\sqrt{2g(u_i, \delta)} = \Delta_i - \sqrt{2g(u_i, \delta)}$})\\
    &\leq \1 \cdot \delta + 0 \cdot (1+\delta) = \delta
\end{align*}

Hence 
$$
\EE[n_i(T)] = \EE[n_i(T)\1_{G_i}] + \EE[n_i(T)\1_{\hat{G}_i}] \leq u_i + \delta T (T+1)
$$

Assuming $\delta = \frac{1}{T(T+1)}$ and taking $u_i = g^{-1}\left (\frac{\Delta_i^2}{8}, \delta \right )$, where $g^{-1}$ is such that $g\left (g^{-1}(x, \delta), \delta \right ) = x$, we get

$$
\EE[n_i(T)] \leq g^{-1}\left (\frac{\Delta_i^2}{8}, \frac{1}{T(T+1)} \right ) + 1
$$

Now, we can proceed with the regret estimation.

\begin{align*}
    &R_T = \sum_{i: \Delta_i>0} \Delta_i \EE[n_i(T)] \leq \sum_{i: \Delta_i>0} \Delta_i \left (g^{-1}\left (\frac{\Delta_i^2}{8}, \frac{1}{T(T+1)} \right ) + 1 \right )
\end{align*}

\endproof

\section{Proof of Theorem~\ref{thm:ZO_UCB}}
\label{sec::proof-ZO-UCB}
Here we provide the detailed proof of Theorem~\ref{thm:ZO_UCB} and its formulation for the reader's convenience.

\begin{theorem}[Convergence of \algname{ZO-UCB}]
 The regret of the \algname{ZO-UCB} with $g(k, \delta)$-bounded first-order algorithm for the MAB problem with $K$ arms, auxiliary functions $f_i(x) = |x-\mu_i|$, period $T$, $\delta=\frac{1}{T(T+1)}$ satisfies
$$
R_T = \sum_{i=1}^K \Delta_i \left ( g^{-1}\left (\frac{\Delta_i}{2}, \frac{1}{T(T+1)} \right ) + 1\right )
$$
\end{theorem}

\proof

The general scheme for the proof is the same as that of Theorem~\ref{thm:FO_UCB}. 
The main differences are highlighted.

First, the complement event is
$$
\hat{G}_i = \left \{\mu_{i^*} > \min_{1 \leq t \leq T} UCB(i^*, n_{i^*}(t), \delta, t)\right \} \cup \left \{x_i^{u_i} + g(u_i, \delta)>\mu_{i^*} \right \}.
$$

with bound on the probability of the second part
\begin{align*}
    &\mathbb{P} \left [x_i^{u_i} + g(u_i, \delta)>\mu_{i^*} \right ]\\
    &=\mathbb{P} \left [x_i^{u_i} - \mu_i + g(u_i, \delta)>\mu_{i^*} - \mu_i \right ] \\
    &=\mathbb{P} \left [x_i^{u_i} - \mu_i > \Delta_i - g(u_i, \delta) \right ]\\
    &=\mathbb{P} \left [x_i^{u_i} - \mu_i > \Delta_i - g(u_i, \delta) | \quad |x_i^{u_i} - \mu_i |> g(u_i, \delta) \right ] \cdot \mathbb{P} \left [ |x_i^{u_i} - \mu_i| > g(u_i, \delta) \right ]\\
    &+\mathbb{P} \left [x_i^{u_i} - \mu_i > \Delta_i - g(u_i, \delta) | \quad |x_i^{u_i} - \mu_i| \leq  g(u_i, \delta) \right ] \cdot \mathbb{P} \left [ |x_i^{u_i} - \mu_i |\leq g(u_i, \delta) \right ]\\
    &(\text{choose $u_i$: $g(u_i, \delta) = \Delta_i - g(u_i, \delta)$})\\
    &\leq \1 \cdot \delta + 0 \cdot (1+\delta) = \delta.
\end{align*}
where we get the same estimation of probability as in the proof of the Theorem \ref{thm:FO_UCB}, but with $u_i = g^{-1}\left (\frac{\Delta_i}{2}, \delta \right )$, which results in

$$
\EE[n_i(T)] \leq g^{-1}\left (\frac{\Delta_i}{2}, \frac{1}{T(T+1)} \right ) + 1
$$
and corresponding regret
$$
R_T = \sum_{i=1}^K \Delta_i \left ( g^{-1}\left (\frac{\Delta_i}{2}, \frac{1}{T(T+1)} \right ) + 1\right ),
$$

\endproof

\section{Proof of Theorem~\ref{th::clip-sgd-th}}
\label{sec::proof-clip-sgd-th}
In this section, we present the proof of Theorem~\ref{th::clip-sgd-th} and provide its formulation for the reader's convenience.
\setcounter{theorem}{7}
\begin{theorem}
    Consider the problem, where $f(x) = \frac12 (x - \mu)^2$ that is 1-strongly convex, satisfies Assumption~\ref{as:L_smoothness}, and the oracle gives an unbiased gradient estimate.
    Also, we assume that the noise in the gradient estimate satisfies Assumption~\ref{as:bounded_bias_and_variance}.
    Then, there exists $C > 0$ such that the \algname{clipped-SGD} with learning rate $\gamma = \min \left( \frac{1}{400L \ln \frac{4(K+1)}{\delta}},  \frac{\ln ((K+1)R^2)}{K+1} \right)$ and clipping hyperparameter $\lambda_k = \frac{\exp(-\gamma (1 + k/2)) R}{120 \gamma \ln \frac{4(K+1)}{\delta}}$ provides the iterates such that after $k = 1,\ldots,K$ iterations the following bound holds with probability at least $1 - \delta$
    \[
    f(x_k) - f^* \leq  C \frac{\ln \frac{4(K+1)}{\delta} \ln^2 ((K+1)R^2)}{k+1}
    \]
    where $K$ is sufficiently large and $R \geq \|x_0 - x^*\|$.
\end{theorem}

\proof

The proof is based on the following theorem from~\cite{puchkin2023breaking}.

\begin{theorem}
\label{thm:clipped_SGD_str_cvx_appendix}
    Let Assumptions~\ref{as:L_smoothness} and strongly convexity property with $\mu > 0$ hold on $Q = B_{2R}(x^*)$, where $R \geq \|x^0 - x^*\|$. Assume that $\nabla f_{\Xi^k}(x^k)$ satisfies Assumption~\ref{as:bounded_bias_and_variance} with parameters $b_k, \sigma_k$ for $k = 0,1,\ldots,K$, $K > 0$ and
    \begin{eqnarray}
        0< \gamma &\leq& \min\left\{\frac{1}{400 L\ln \tfrac{4(K+1)}{\delta}}, \frac{\ln(B_K)}{\mu(K+1)}, \frac{\ln(C_K)}{\mu(1+ \nicefrac{K}{2})}, \frac{2\ln(D)}{\mu (K+1)}\right\}, \label{eq:gamma_SGDA_str_mon}\\
        B_K &=& \max\left\{2, \frac{(K+1)\mu^2R^2}{5400\sigma^2\ln\left(\frac{4(K+1)}{\delta}\right)\ln^2(B_K)} \right\} = \cO\!\left(\!\max\!\left\{2, \frac{K\mu^2R^2}{\sigma^2\ln\left(\!\frac{K}{\delta}\!\right)\ln^2\left(\!\max\!\left\{2, \frac{K\mu^2R^2}{\sigma^2\ln\left(\!\frac{K}{\delta}\!\right)} \right\}\right)} \right\}\!\right), \label{eq:B_K_SGD_str_cvx_2} \\
        C_K &=& \max\left\{2, \frac{(\frac{K}{2}+1)\mu R}{480 b\ln\left(\frac{4(K+1)}{\delta}\right)\ln(C_K)} \right\} = \cO\left(\max\left\{2, \frac{K\mu R}{b\ln\left(\frac{K}{\delta}\right)\ln\left(\max\left\{2, \frac{K\mu R}{b\ln\left(\frac{K}{\delta}\right)} \right\}\right)} \right\}\right), \label{eq:C_K_SGD_str_cvx_2} \\
        D &=& \max\left\{2, \frac{\mu R}{80 b\ln(D)} \right\} = \cO\left(\max\left\{2, \frac{\mu R}{b\ln\left(\max\left\{2, \frac{\mu R}{b} \right\}\right)} \right\}\right), \label{eq:D_K_SGD_str_cvx_2} \\
        \lambda_k &=& \frac{\exp(-\gamma\mu(1 + \nicefrac{k}{2}))R}{120\gamma \ln \tfrac{4(K+1)}{\delta}}, \label{eq:lambda_SGDA_str_mon}
    \end{eqnarray}
    for some $\delta \in (0,1]$ and $b = \max_{k=0,1,\ldots,K} b_k$, $\sigma = \max_{k=0,1,\ldots,K} \sigma_k$. Then, after $K$ iterations the iterates produced by \algname{clipped-SGD} with probability at least $1 - \delta$ satisfy 
    \begin{equation}
        \|x^{K+1} - x^*\|^2 \leq 2\exp(-\gamma\mu(K+1))R^2. \label{eq:main_result_str_cvx_SGD_appendix}
    \end{equation}
    In particular, when $\gamma$ equals the minimum from \eqref{eq:gamma_SGDA_str_mon}, then the iterates produced by \algname{clipped-SGD} after $K$ iterations with probability at least $1-\delta$ satisfy
    \begin{equation}
       \|x^{K} - x^*\|^2 = \cO\left(\max\left\{R^2\exp\left(- \frac{\mu K}{L \ln \tfrac{K}{\delta}}\right), \frac{\sigma^2\ln\left(\frac{K}{\delta}\right)\ln^2\left(B_K\right)}{K\mu^2}, \frac{bR\ln\left(\frac{K}{\delta}\right)\ln\left(C_K\right)}{K\mu}, \frac{bR\ln(D)}{\mu}\right\}\right). \label{eq:clipped_SGD_str_cvx_appendix}
    \end{equation}
\end{theorem}

In this theorem we take the symmetric noise, therefore we ignore terms $C_K$ and $D$.
Also, we consider the sufficiently large number of iterations that
$\exp\left({-\frac{K+1}{400L \ln \frac{4(K+1)}{\delta}}}\right) < \frac{C \ln \frac{4(K+1)}{\delta} \ln^2 B_K}{(K+1)} $ holds, where $C = 5400\sigma^2$.
To get the proper bounds for $B_K$ related terms we use the following expression $\ln^2 B_k \leq \ln^2 \frac{(K+1)R^2}{C \ln^2 \frac{4(K+1)}{\delta}} \leq \ln^2 ((K+1)R^2)$ obtained from the simple estimate $(x - y)^2 \leq \max (x^2, y^2)$.

\endproof

\section{Proof of Theorem~\ref{thm:MAB_theorem}}
\label{sec::proof-MAB-theorem}

Here we provide the detailed proof of Theorem~\ref{thm:MAB_theorem} and its formulation for the reader's convenience.

\begin{theorem}[Convergence of \algname{Clipped-SGD-UCB}]
 The regret of the \algname{Clipped-SGD-UCB} for multi-armed bandit problem with $K$ arms, period $T$, $\gamma_t = $, $\lambda_k=$ and symmetric distribution of rewards satisfies:
 $$
R_T \leq 4\log((T+1)TR^2)\sqrt{C \log(4T(T+1)^2)TK} +  \sum_i \Delta_i $$
$$
R_T \leq \sum_{i: \Delta_i>0}  \left [ \Delta_i + \frac{8C \log(4T(T+1^2)) \log^2((T+1)TR^2)}{\Delta_i}\right ] 
$$

\end{theorem}

\proof

From Theorem \ref{thm:FO_UCB} we get
\begin{align*}
    &R_T \leq \sum_{i: \Delta_i>0} \Delta_i \left (g^{-1}\left (\frac{\Delta_i^2}{8}, \frac{1}{T(T+1)} \right ) + 1 \right )
\end{align*}

In case $g(k, \delta) = \frac{C \log(4(T+1)/\delta) \log^2((T+1)R^2)}{k}$ we get $g^{-1}(x, \delta) = \frac{C \log(4(T+1)/\delta) \log^2(T+1)R^2}{x}$, and

$$
R_T \leq \sum_{i: \Delta_i>0}  \left [ \Delta_i + \frac{8C \log(4T(T+1^2)) \log^2((T+1)TR^2)}{\Delta_i}\right ] 
$$

We get instance-dependent bound. Now let $\Delta>0$ be some fixed value. Then we can bound regret in the following way:

\begin{align*}
    &R_T = \sum_{i: \Delta_i>0} \Delta_i \EE[n_i(T)] = \sum_{i: \Delta_i<\Delta} \Delta_i \EE[n_i(T)] + \sum_{i: \Delta_i\geq \Delta} \Delta_i \EE[n_i(T)]  \\
    &\leq T \Delta + \frac{8 C \log(4T(T+1)^2) \log^2((T+1)TR^2) K \log T}{\Delta} +  \sum_i \Delta_i \\ 
    &\left (\text{choose $\Delta = \frac{ \log((T+1)TR^2)\sqrt{8C \log(4T(T+1)^2) K}}{\sqrt{T}}$} \right )\\
    &\leq 4\log((T+1)TR^2)\sqrt{C \log(4T(T+1)^2)TK} +  \sum_i \Delta_i
\end{align*}

\endproof

\section{Additional experiments for super heavy tail MAB}
\label{sec::app_exp_heavy}

\begin{figure}[!h]
    \centering
    \begin{subfigure}{0.3\textwidth}
    \includegraphics[width=\linewidth]{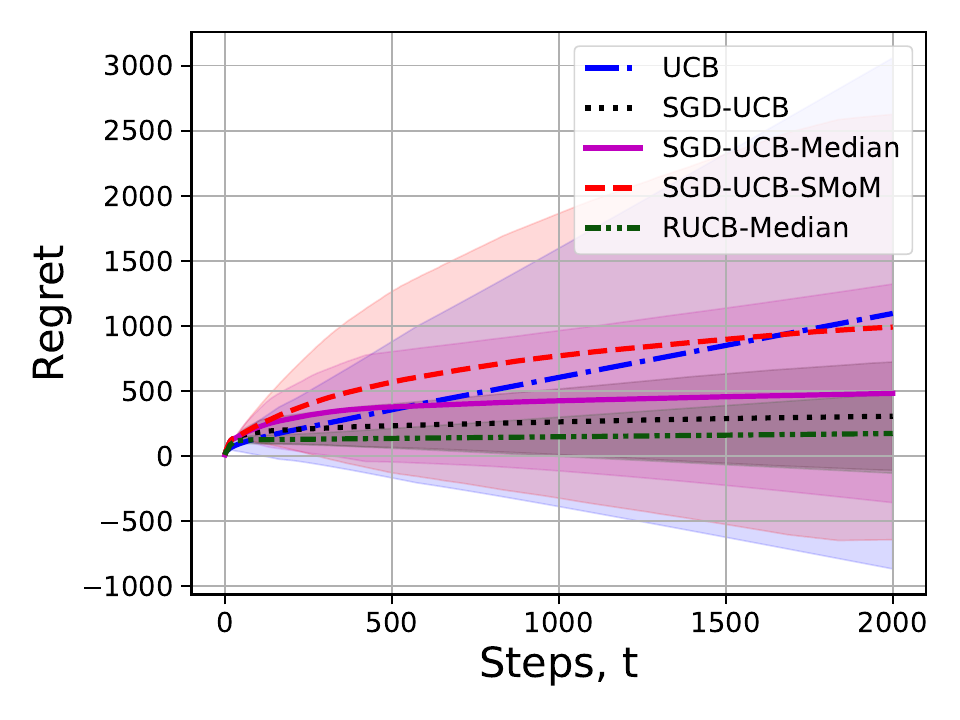}
        \caption{\texttt{Env1}}
    \end{subfigure}
    ~
    \begin{subfigure}{0.3\textwidth}
    \includegraphics[width=\linewidth]{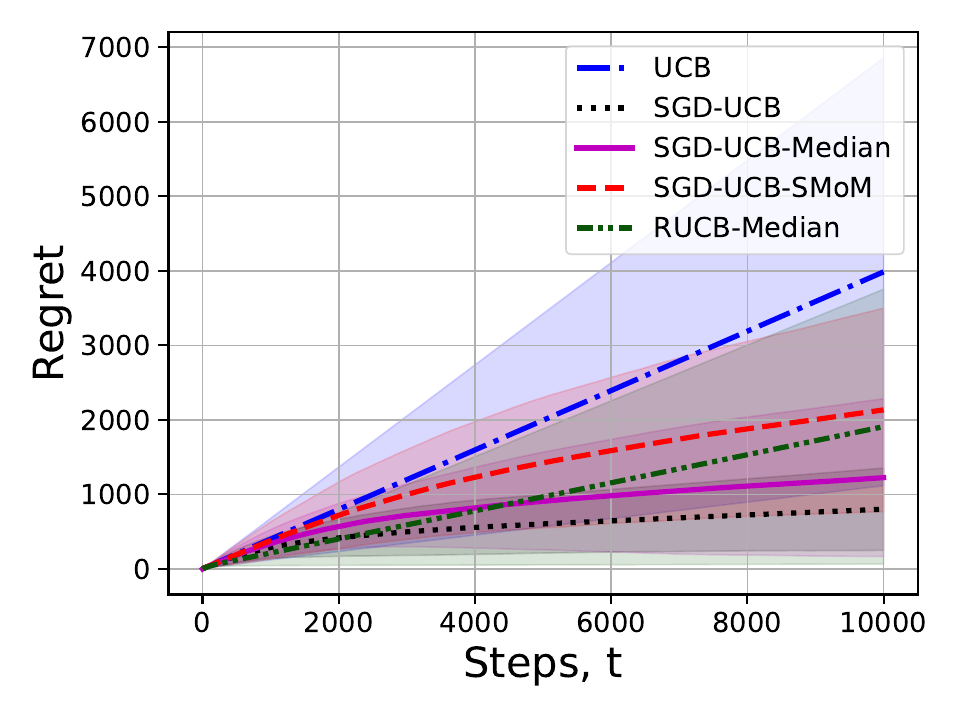}
        \caption{\texttt{Env2}}
    \end{subfigure}
    ~
    \begin{subfigure}{0.3\textwidth}
    \includegraphics[width=\linewidth]{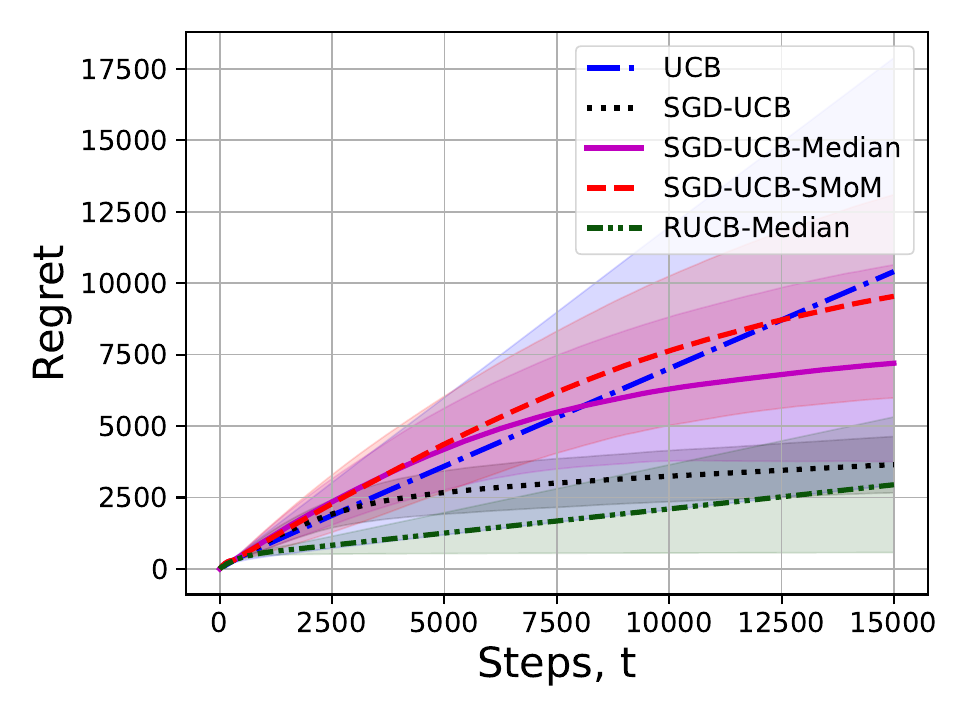}
        \caption{\texttt{Env3}}
    \end{subfigure}
    \\
    \begin{subfigure}{0.3\textwidth}
    \includegraphics[width=\linewidth]{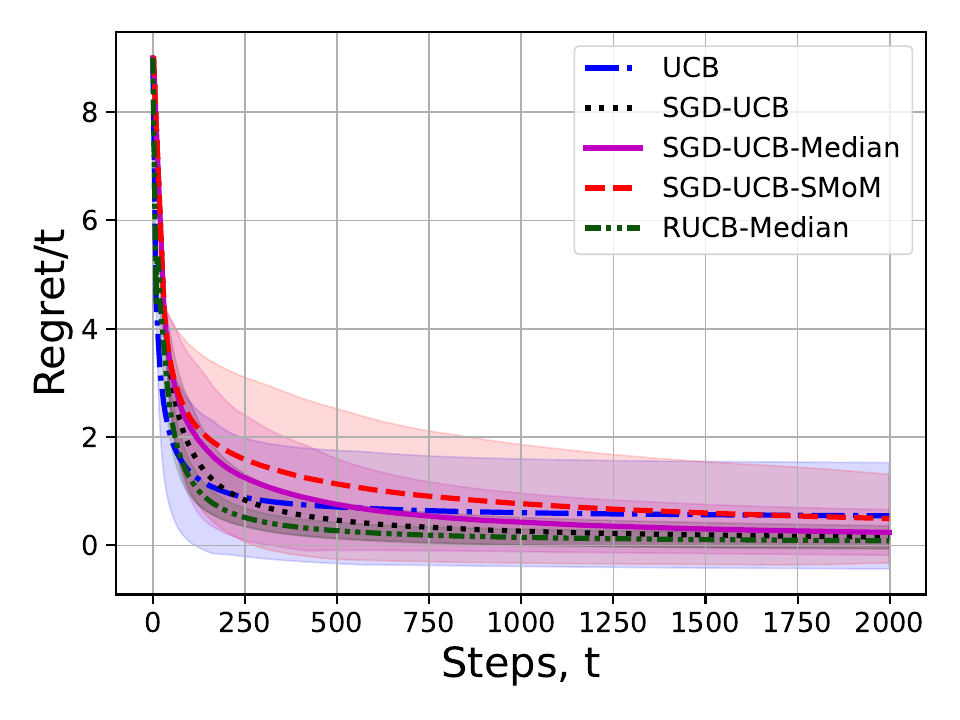}
        \caption{\texttt{Env1}}
    \end{subfigure}
    ~
    \begin{subfigure}{0.3\textwidth}
    \includegraphics[width=\linewidth]{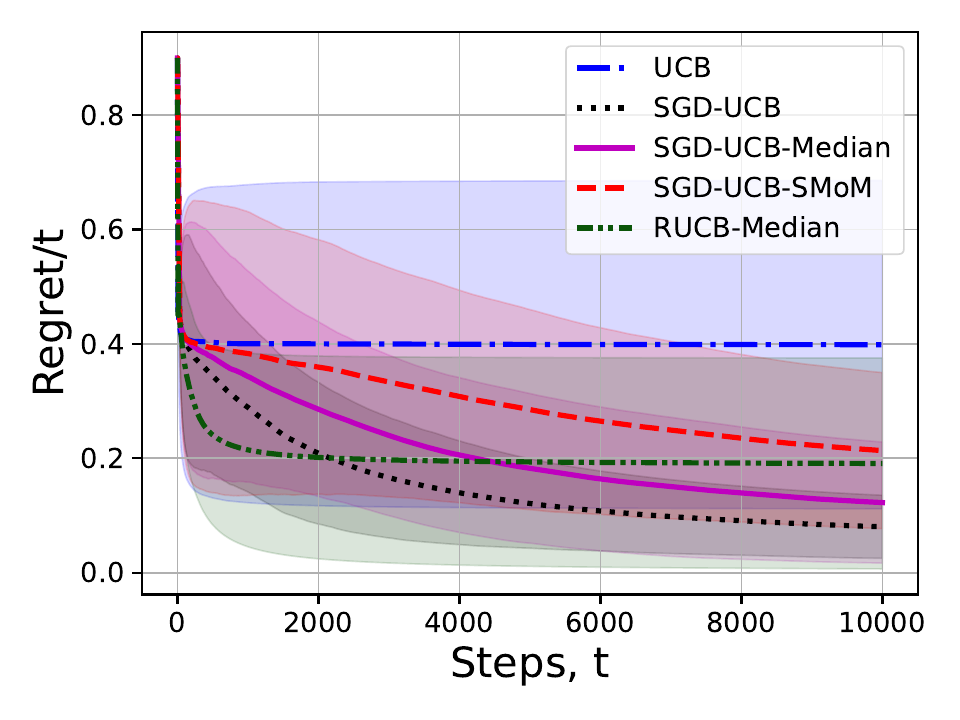}
        \caption{\texttt{Env2}}
    \end{subfigure}
    ~
    \begin{subfigure}{0.3\textwidth}
    \includegraphics[width=\linewidth]{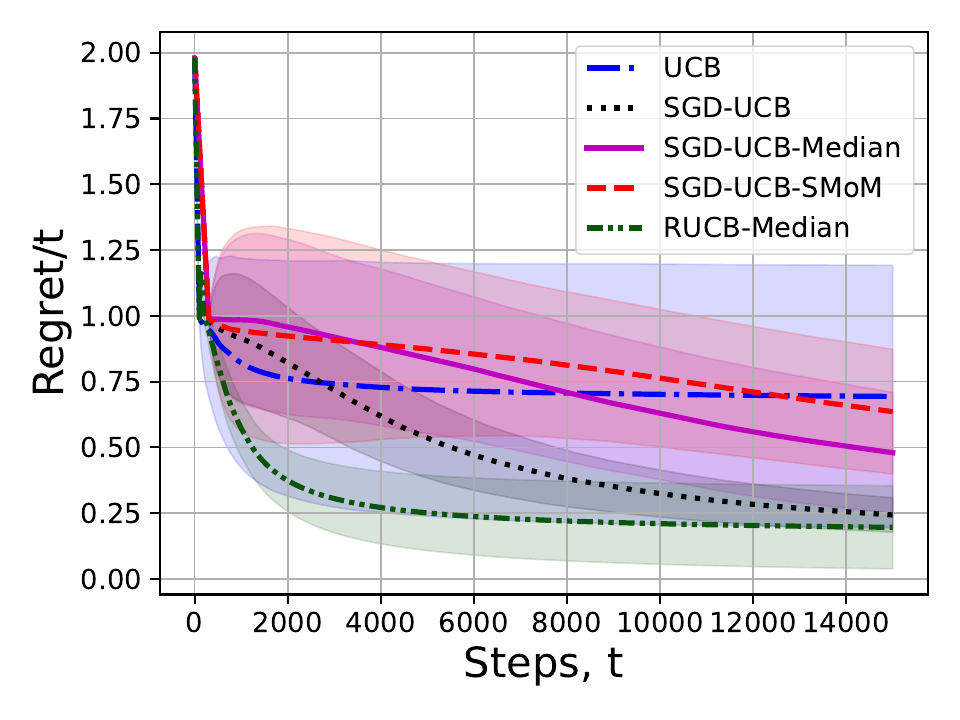}
        \caption{\texttt{Env3}}
    \end{subfigure}
    \caption{Regret and mean regret comparison for reward noise generated from Fr\'echet ($\alpha=1$) distribution.}
    \label{fig::frechet_1}
\end{figure}

\begin{figure}[!h]
    \centering
    \begin{subfigure}{0.3\linewidth}
    \includegraphics[width=\textwidth]{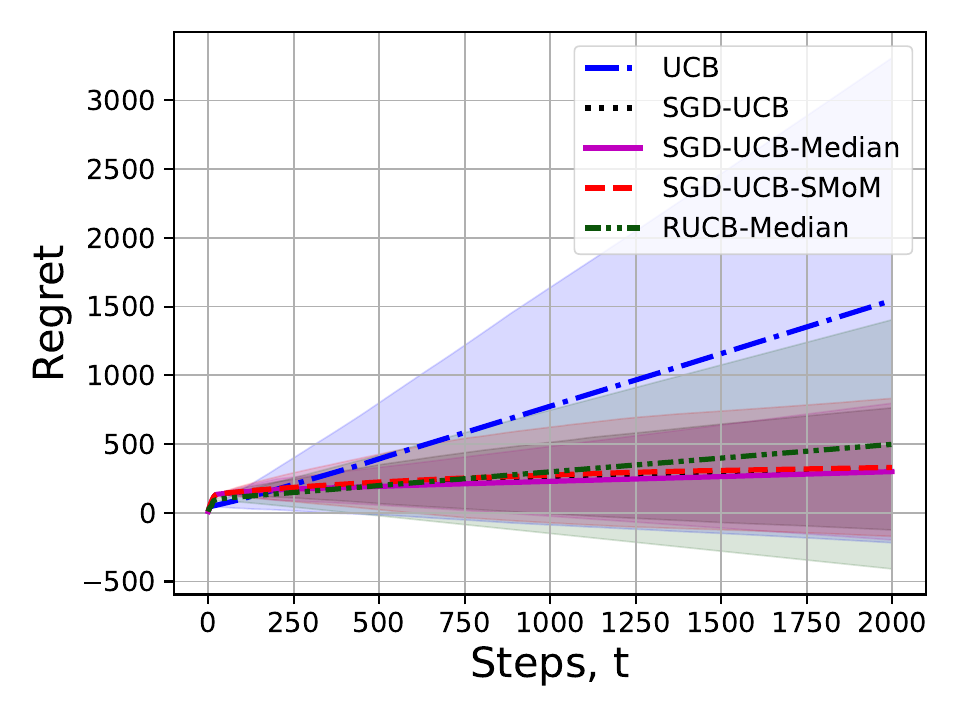}
        \caption{\texttt{Env1}}
    \end{subfigure}
    ~
    \begin{subfigure}{0.3\linewidth}
    \includegraphics[width=\linewidth]{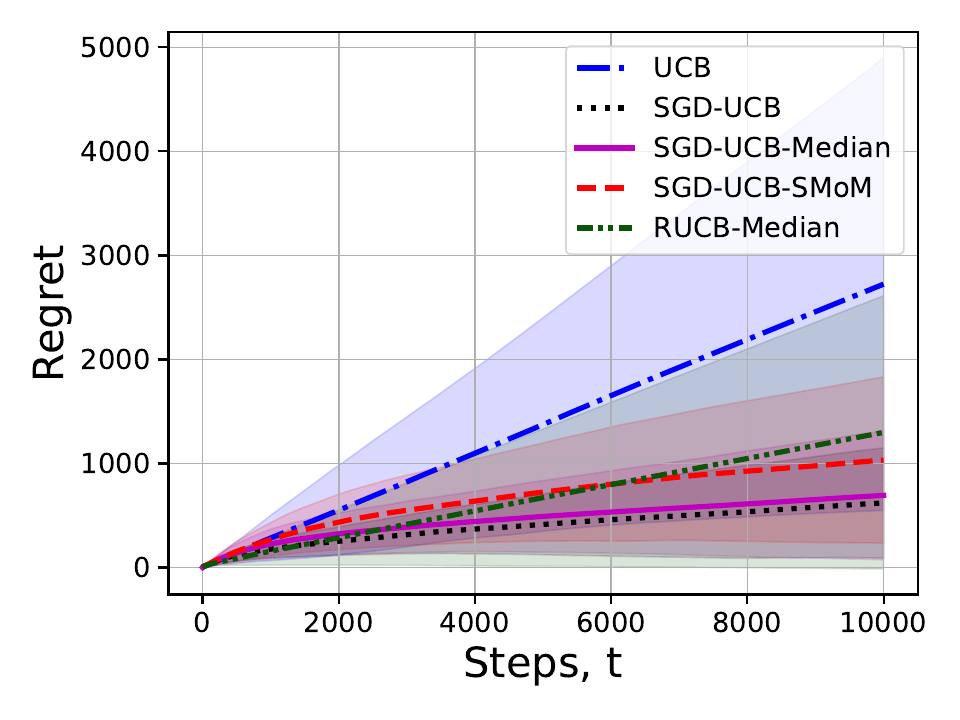}
        \caption{\texttt{Env2}}
    \end{subfigure}
    ~
    \begin{subfigure}{0.3\linewidth}
    \includegraphics[width=\linewidth]{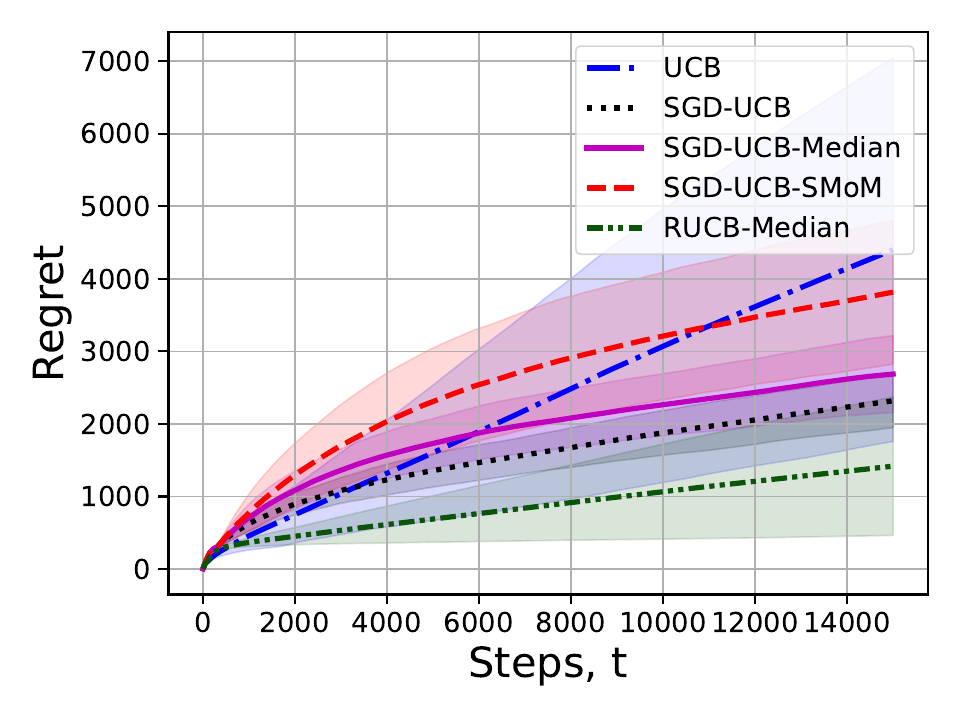}
        \caption{\texttt{Env3}}
    \end{subfigure}
    \\
    \begin{subfigure}{0.3\linewidth}
    \includegraphics[width=\linewidth]{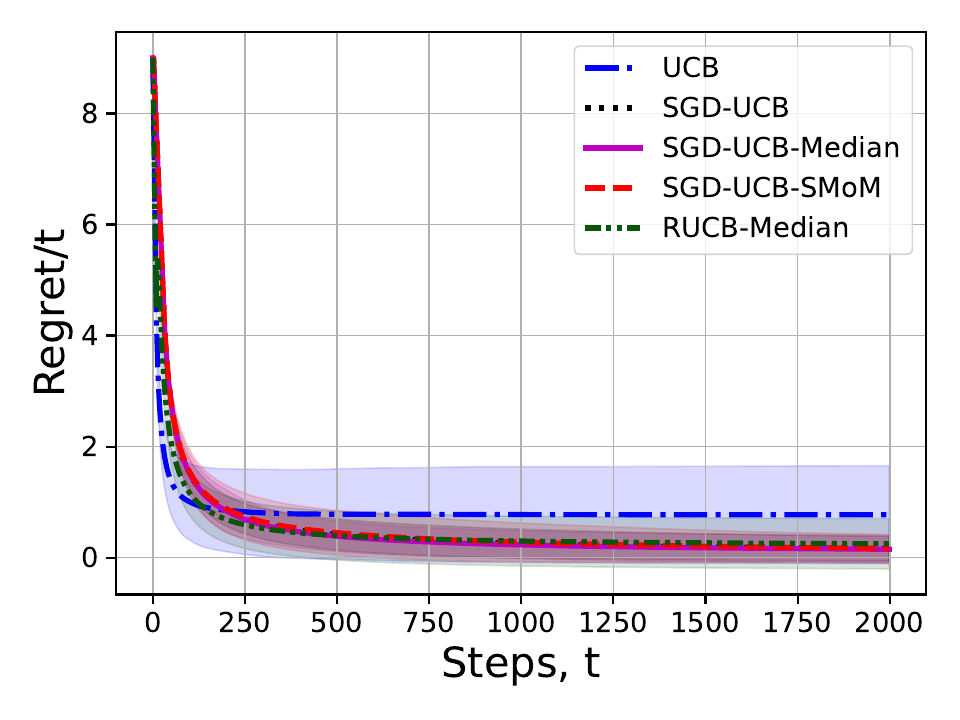}
        \caption{\texttt{Env1}}
    \end{subfigure}
    ~
    \begin{subfigure}{0.3\linewidth}
    \includegraphics[width=\linewidth]{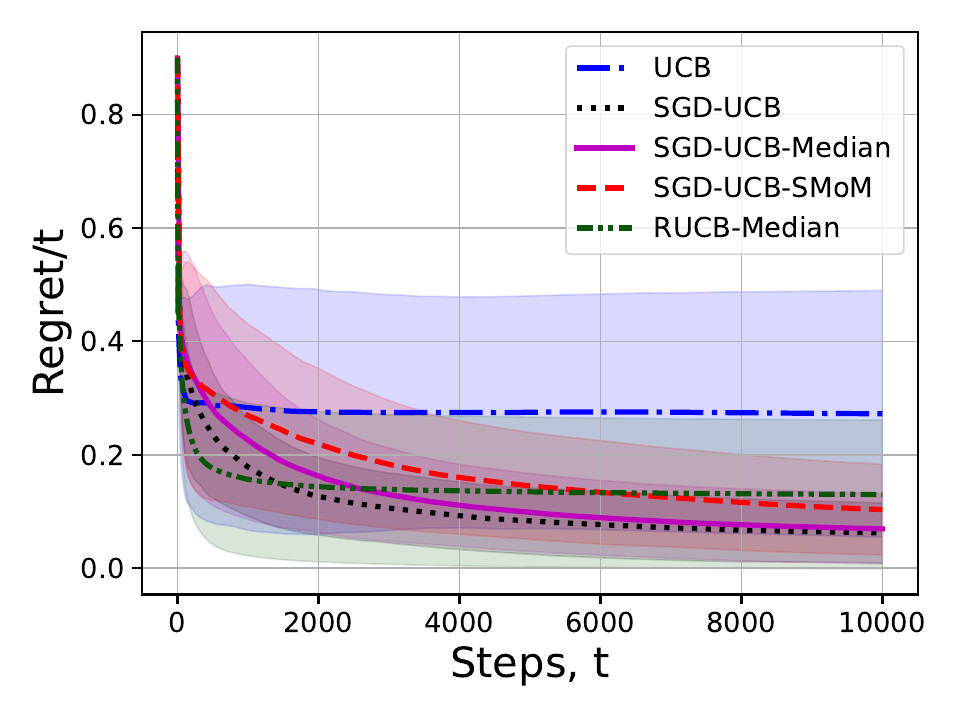}
        \caption{\texttt{Env2}}
    \end{subfigure}
    ~
    \begin{subfigure}{0.3\linewidth}
    \includegraphics[width=\linewidth]{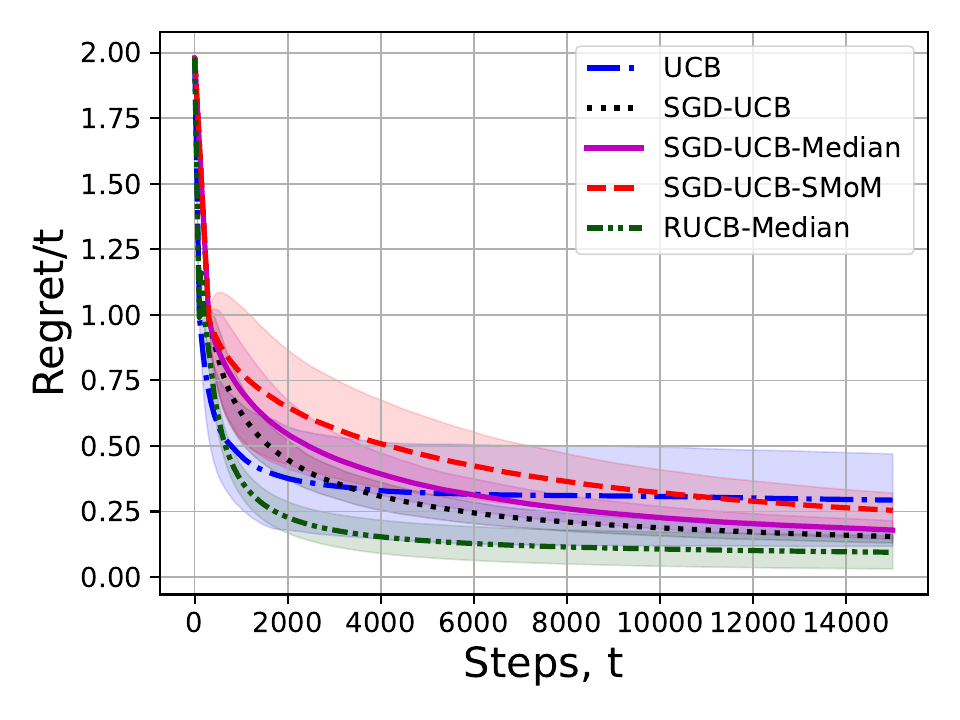}
        \caption{\texttt{Env3}}
    \end{subfigure}
    \caption{Regret and mean regret comparison for reward noise generated from the mixture Cauchy $(\gamma = 1)$ and exponential distributions with weights $0.7$ and $0.3$, respectively.}
    \label{fig::cauchyexp}
\end{figure}


\begin{figure}[!h]
    \centering
    \begin{subfigure}{0.3\linewidth}
    \includegraphics[width=\linewidth]{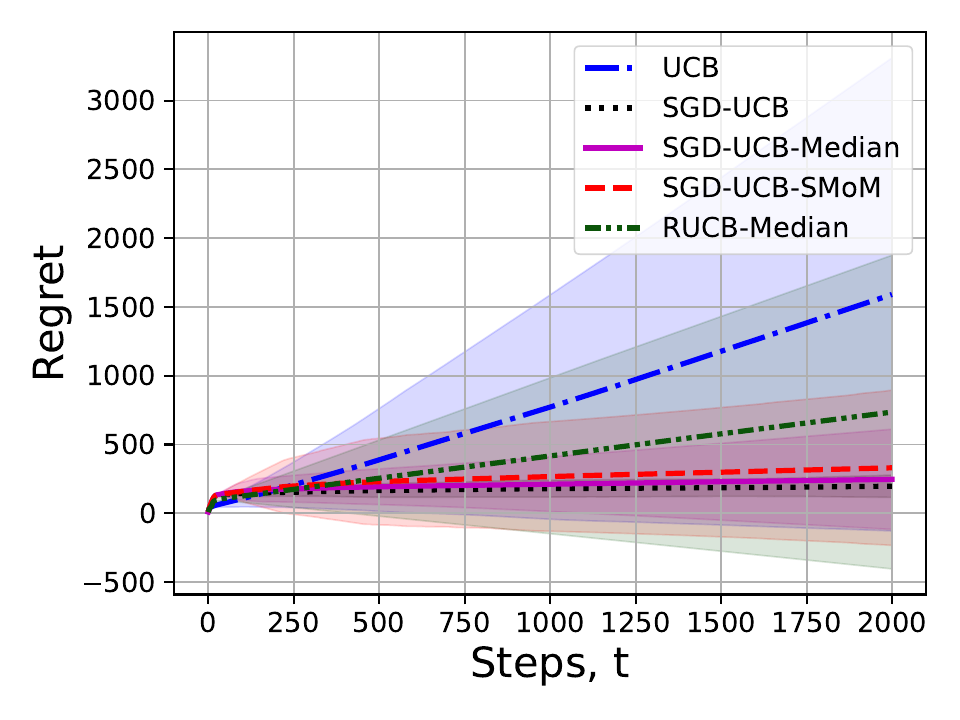}
        \caption{\texttt{Env1}}
    \end{subfigure}
    ~
    \begin{subfigure}{0.3\linewidth}
    \includegraphics[width=\linewidth]{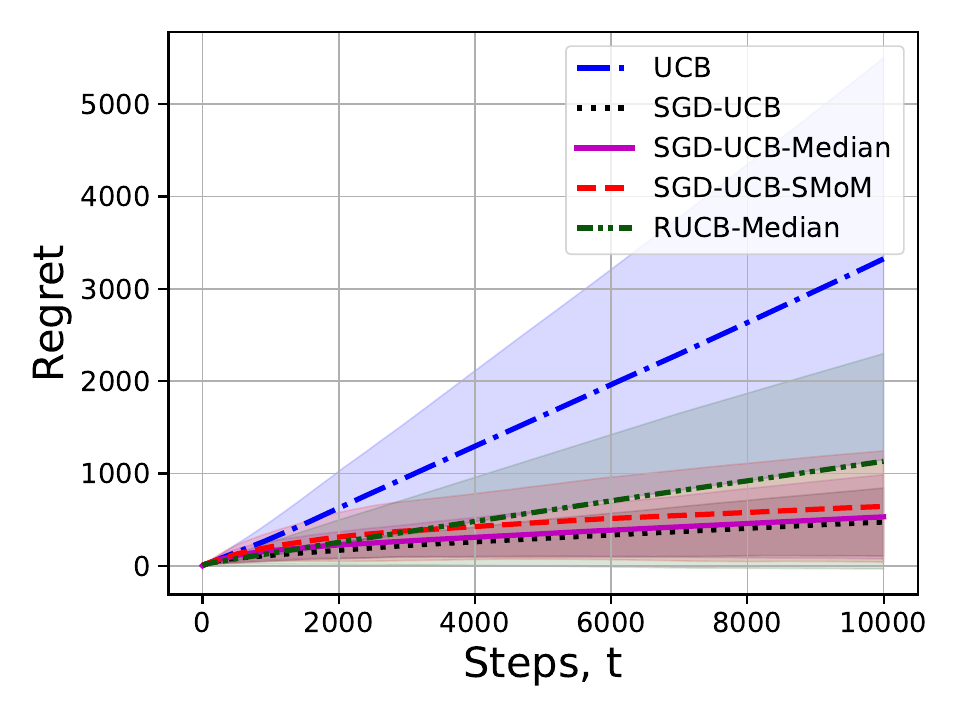}
        \caption{\texttt{Env2}}
    \end{subfigure}
    ~
    \begin{subfigure}{0.3\linewidth}
    \includegraphics[width=\linewidth]{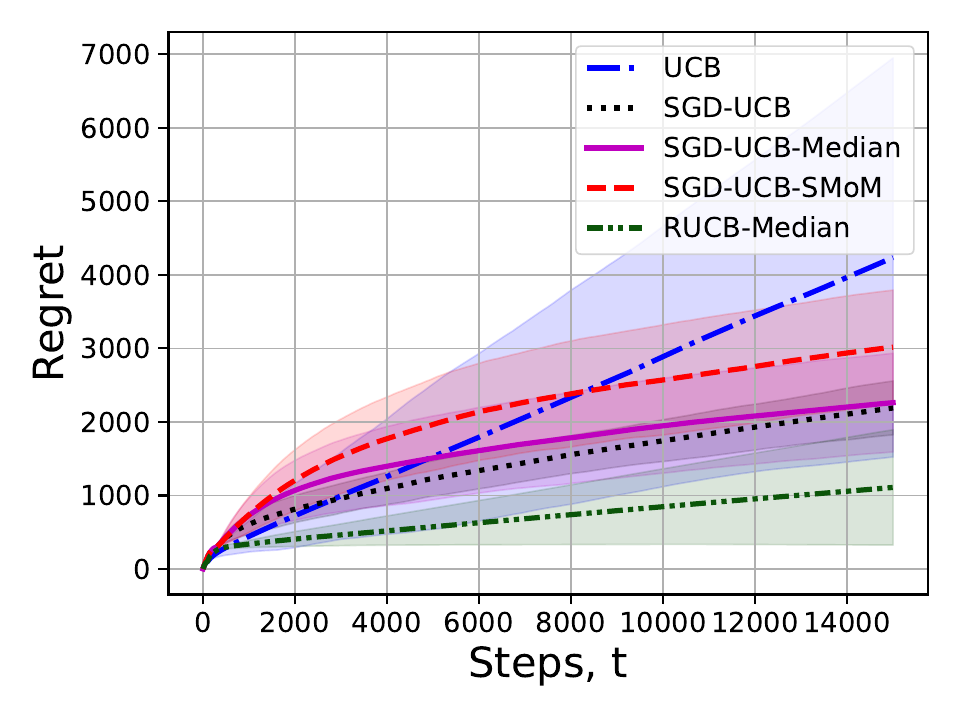}
        \caption{\texttt{Env3}}
    \end{subfigure}
    \\
    \begin{subfigure}{0.3\linewidth}
    \includegraphics[width=\linewidth]{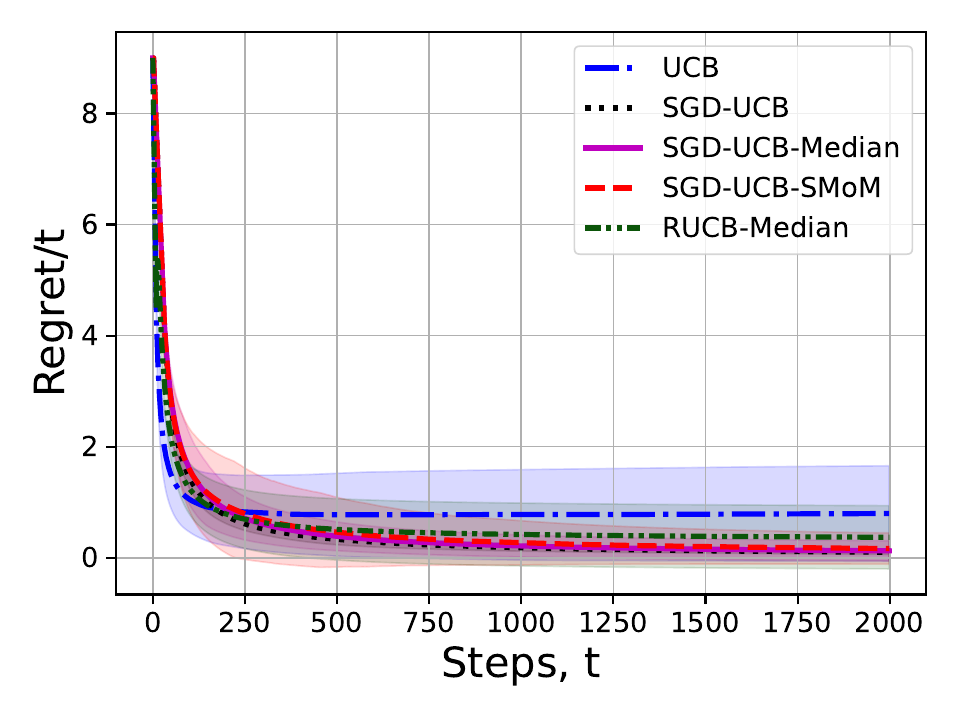}
        \caption{\texttt{Env1}}
    \end{subfigure}
    ~
    \begin{subfigure}{0.3\linewidth}
    \includegraphics[width=\linewidth]{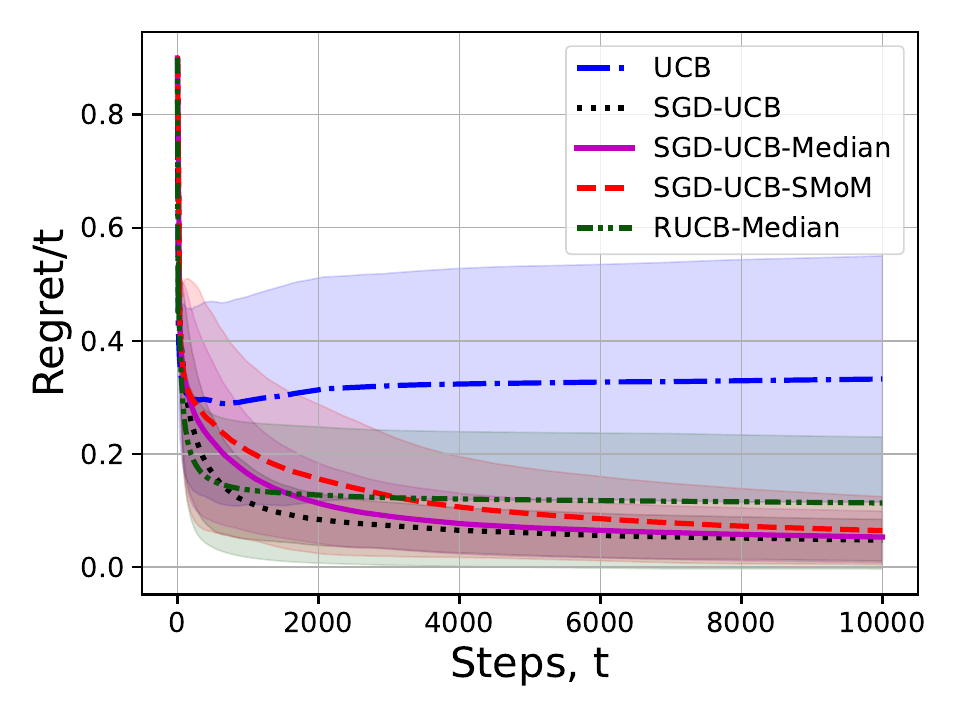}
        \caption{\texttt{Env2}}
    \end{subfigure}
    ~
    \begin{subfigure}{0.3\linewidth}
    \includegraphics[width=\linewidth]{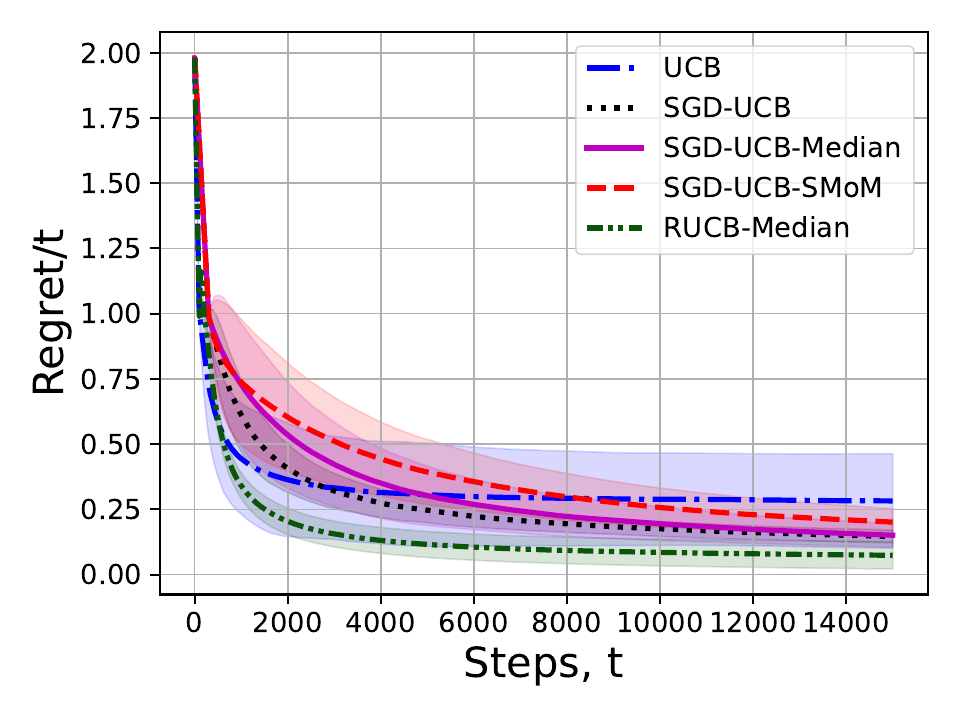}
        \caption{\texttt{Env3}}
    \end{subfigure}
    \caption{Regret and mean regret comparison for reward noise generated from the mixture Cauchy $(\gamma = 1)$ and Pareto distributions with weights $0.7$ and $0.3$, respectively.}
    \label{fig::cauchypareto}
\end{figure}

\end{document}